%% file: main.tex
\documentclass{article}
\usepackage{aikyam}						

\usepackage{booktabs}						
\usepackage{multirow}						
\usepackage{amsfonts}						
\usepackage{graphicx}						
\usepackage{duckuments}						
\usepackage[most]{tcolorbox}
\usepackage{lineno}

\usepackage[numbers,compress,sort]{natbib}	

\title[\method: Multi-modal Explanation Agent]{\method: A Reward-Driven Multi-Agent System for Faithful Model Explanations}

\author[Cheng et al.]{%
Yuyang Cheng$^{1}$\thanks{Corresponding author: \href{mailto:jrm9ga@virginia.edu}{Yuyang Cheng}}
\And
Raghav Kaushik Ravi$^{2}$
\And
Srivarshinee Sridhar$^{2}$
\And
Sriparna Saha$^{3}$
\And
Akash Ghosh$^{3}$
\And
Chirag Agarwal$^{1}$
}
\newcommand{\pvar}[1]{\textcolor{violet!85!black}{$\langle$\textit{#1}$\rangle$}}

\newcommand{\xhdr}[1]{\vspace{0em}\noindent{{\bf #1.}}}
\newcommand{\ie}{\textit{i.e., \xspace}}
\newcommand{\eg}{\textit{e.g., \xspace}}
\newcommand{\method}{\textsc{Mea}\xspace}

\newcommand{\methodbench}{\textsc{Mea-Bench}\xspace}
\newcommand{\hide}[1]{}

\definecolor{greenblue}{RGB}{0,150,136}

\usepackage{colortbl}

\definecolor{Gray}{gray}{0.9}
\definecolor{LightCyan}{rgb}{0.88,1,1}
\definecolor{darkred}{rgb}{0.8,0.1,0.1}
\definecolor{darkyellow}{rgb}{0.95, 0.68, 0.22}
\definecolor{darkgreen}{rgb}{0.1,0.8,0.1}
\newcolumntype{a}{>{\columncolor{Gray}}c}
\newcolumntype{b}{>{\columncolor{white}}c}

\begin{document}

\maketitle
\begin{center}
    \small
    \vspace{-0.25in}
    $^1$~University of Virginia~~~$^2$~Vellore Institute of Technology~~~$^3$~Indian Institute of Technology, Patna
    
    \vspace{0.5em}
    \textbf{Project Page:} \url{https://estherrcheng.github.io/mea_website/} \\
    \textbf{Code:} \url{https://github.com/AikyamLab/xai-agent}
\end{center}
\input{000abstract}

\input{010intro}

\input{020related}
\input{030method}
\newpage
\input{040results}
\input{050conclusion}

\section*{Acknowledgements}
We would like to thank all members of \href{https://chirag-agarwall.github.io/}{Aikyam Lab} for their valuable feedback. C.A. is supported, in part, by grants from Capital One, LaCross Institute for Ethical AI in Business, the UVA Environmental Institute, OpenAI Researcher Program, Thinking Machine's Tinker Research Grant, and Cohere. The views expressed are those of the authors and do not reflect the official policy or the position of the funding agencies.

\bibliographystyle{unsrtnat}
\bibliography{reference}

\newpage

\appendix

\input{060appendix}

\end{document}

%% file: 000abstract.tex
\begin{abstract}
\looseness=-1 Recent years have seen the employment of a plethora of machine learning (ML) models in high-stakes domains, but they remain largely opaque to the practitioners who act on their predictions. While post-hoc explanation methods offer a lens into this model behavior, wielding them effectively demands expertise most domain experts lack: navigating high-dimensional outputs, selecting the best explanations, and synthesizing evidence across disparate tools. To this end, we present \method, a multi-agent framework that removes the explanation knowledge barrier entirely: a \textit{Proposer} agent selects and configures explanation tools based on the question and modality, while an \textit{Actor} agent is optimized end-to-end against faithfulness, transforming the outputs into natural language explanations grounded in model behavior across tabular, text, and vision modalities. Further, we introduce diverse question types spanning feature attribution, counterfactual reasoning, and spurious feature detection, each paired with a perturbation-based faithfulness metric. We find that frontier 
LLMs systematically produce unfaithful explanations. By optimizing against faithfulness rewards augmented with a modality-adaptive penalty, \method consistently outperforms post hoc explainers, agentic, and closed-source baselines across six datasets, with reward-driven optimization yielding faithfulness gains of $+28\%$ (tabular), $+21\%$ (text), and $+34\%$ (vision) over the untrained backbone. More broadly, our findings suggest that AI agents themselves can serve as a scalable, adaptable interface to ML explainability, opening a path toward natural-language explainability that generalizes beyond the fixed, single-purpose tools that have long defined the field.

\end{abstract}

%% file: 010intro.tex
\section{Introduction}
While machine learning (ML) models are increasingly deployed in high-stakes domains, such as healthcare, finance, and legal decision-making \cite{rajpurkar2022ai,cary2023mitigating,cao2024fundamental,flores2016false}, their opacity has become a critical concern for practitioners, regulators, and end users, hindering their trustworthiness. This gap between predictive performance and human interpretability is a central challenge that explainability research seeks to address. More broadly, foundation models can generate hallucinated or unsupported outputs, making reliability a particularly important concern in high-stakes applications \citep{sahoo2024comprehensive,ghosh2025clinic}.

\looseness=-1 Prior works have introduced a plethora of post-hoc explanation methods~\citep{ribeiro2016whyitrustyou,sundararajan2017axiomaticattributiondeepnetworks,Selvaraju_2019, lundberg2017unifiedapproachinterpretingmodel, article, smilkov2017smoothgradremovingnoiseadding, springenberg2015strivingsimplicityconvolutionalnet} that probe trained models to generate explanations. However, each method implements a fixed procedure yielding a single output type, \ie an attribution vector, a heatmap, or a set of super-pixels, independent of what the user actually wants to know. Further, deploying them correctly requires \textit{selecting among competing methods, interpreting high-dimensional outputs, and synthesizing results across tools}, imposing a prohibitive knowledge barrier for domain practitioners without ML expertise.\par

\begin{figure*}
    \centering
    \includegraphics[width=\linewidth]{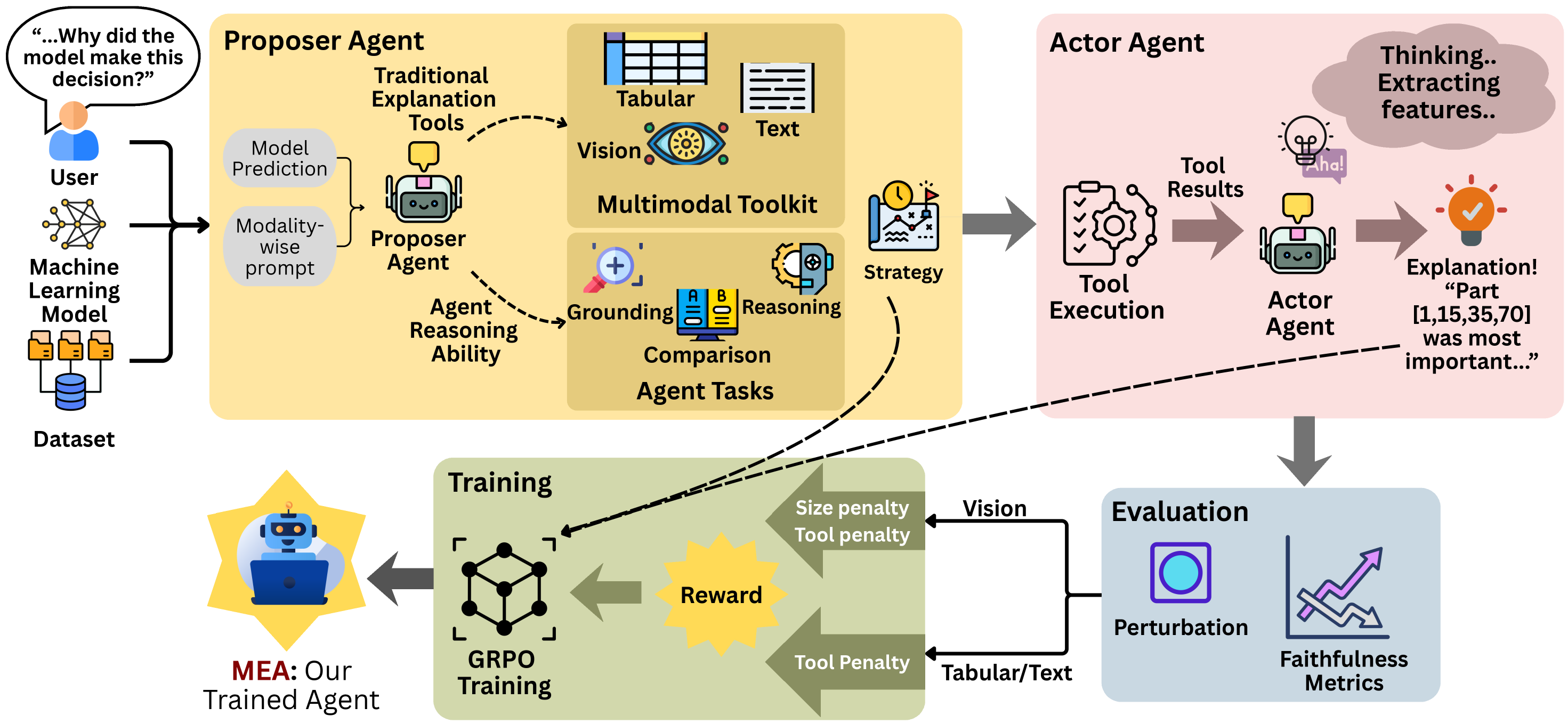}
    \vspace{0.05in}
    \caption{\xhdr{Overview of \method} Given a user question about model behavior, the Proposer agent reasons over the model prediction and input modality to select an explanation strategy, choosing from a multimodal (tabular, text, and vision) toolkit and agent reasoning tasks (grounding, reasoning, and comparison), the Actor agent then executes the strategy, processes raw explanation outputs, and produces a structured natural-language explanation. The explanation is evaluated via input perturbation against faithfulness metrics. In training, both agents are jointly optimized end-to-end via GRPO~\cite{shao2024deepseekmathpushinglimitsmathematical}, using a reward combining faithfulness  and tool penalties.
    }
    \label{fig:pipeline_framework}
\end{figure*}

\looseness=-1 Conversational explanation frameworks (CEF) address this barrier by allowing users to interrogate models in natural language~\cite{slack2023talktomodelexplainingmachinelearning,10.1145/3734191,kim2025talktoagenthumancentricexplanationreinforcement, he2026agenticapproachgeneratingxainarratives}, mapping free-form queries to structured explanation operations without requiring users to select or configure individual methods. However, these systems share three limitations that remain unaddressed. {First}, CEF are largely \textit{training-free}: they rely on hand-crafted parsing rules or prompt engineering to route user queries to explanation tools, \textit{without any learned component that adapts to explanation quality or user intent}, resulting in unfaithful explanations. {Second}, CEF have not been evaluated across large-scale datasets across different modalities. \citet{kroeger2024incontextexplainersharnessingllms} found that LLM-generated explanations fall behind post-hoc explainers and are susceptible to alphabetical bias, questioning whether LLMs can genuinely outperform existing explainers. {Third}, existing CEF lack a principled, quantitative evaluation framework: they either assess quality through single explanation types~\citep{li2023mathcalm, Rawal_2025} or focus on simple linear models, where generated explanations are treated as final outputs with no verification step against model behavior.

\looseness=-1\xhdr{Present work} We address the above gaps by introducing \method, a novel multi-agent framework (see Fig.~\ref{fig:pipeline_framework}) for \emph{prediction-level explainability across diverse modalities}. \method comprises of a Proposer agent that selects and configures an appropriate combination of tools given a question and data modality, and an Actor agent that executes the tools and synthesizes a natural-language explanation. To scale across diverse explanation difficulty and data modalities, \method optimizes a multi-modal agent backbone end-to-end via reinforcement learning, with faithfulness scores to directly incentivize explanations that are verifiably grounded in model behavior rather than merely plausible. Since explanation complexity and scale differ fundamentally across modalities, we further introduce a \emph{modality-adaptive reward} that pairs a universal tool-count penalty with modality-specific constraints, preventing reward hacking while encouraging concise, targeted explanations uniformly across all modalities. Further, we introduce the first comprehensive, model- and modality-agnostic question \textbf{taxonomy}, comprising diverse question types that spans feature attribution, counterfactual reasoning, and spurious feature detection with their respective faithfulness metrics to quantify alignment between the agent's explanation and the model's actual behavior via perturbation-based testing. To enable large-scale training and evaluation, we construct \methodbench, a dataset covering \textit{three data modalities (tabular, text, and vision), six model architectures, and six datasets, comprising 13K training and 3.5K test instances}. 

\looseness=-1 Our main contributions are as follows: 1) we present \method, a multi-agent framework that selects, executes, and synthesizes explanation tools into faithful natural-language explanations across tabular, text, and vision modalities; 2) we introduce \methodbench, comprising diverse faithfulness-verified question types, \textbf{12} dataset-model combinations, and 13k training and 3.5k test instances across three modalities; 3) we demonstrate that frontier agents systematically hallucinate on explainability tasks and address this by optimizing \method end-to-end via reinforcement learning, using faithfulness as the reward augmented by a modality-adaptive penalty that  prevents reward hacking across modalities; and 4) our trained \method outperforms post-hoc explainers, closed-source frontier models, and advanced agentic frameworks, yielding average faithfulness gains of \textbf{+28\%} (tabular), \textbf{+21\%}(text), and \textbf{+34\%} (vision) over the untrained backbone.

%% file: 020related.tex
\section{Related Works}

\looseness=-1\xhdr{Post-hoc Explainability} Post-hoc explainers attribute predictions to input features without modifying the model. Representative approaches include LIME~\citep{ribeiro2016whyitrustyou}, which fits a locally linear surrogate around the instance, and SHAP~\citep{lundberg2017unifiedapproachinterpretingmodel}, which provides Shapley-value-based attributions with consistency guarantees. For vision, Grad-CAM~\citep{Selvaraju_2019} and gradient techniques~\citep{sundararajan2017axiomaticattributiondeepnetworks,smilkov2017smoothgradremovingnoiseadding} supply pixel-level attribution signals.

\looseness=-1\noindent\underline{\smash{\textit{\textbf{Research Gap.}}}} Existing faithfulness evaluation~\citep{li2023mathcalm,zhang2025unifiedattributionexplainableai} is conducted \emph{offline} as a post-hoc assessment; no prior work adapts it into an \emph{online training signal} to optimize an agentic framework across a comprehensive, multi-question, multi-modal setup.

\looseness=-1\xhdr{Conversational and Agentic XAI} Early interactive works framed explanations as dialogues: TalkToModel~\citep{slack2023talktomodelexplainingmachinelearning} enables natural-language exploration of model behavior, and \textsc{Echo}~\citep{10.1145/3734191} grounds answers in computed attributions rather than parametric knowledge. \textsc{Fax}~\citep{kim2025towards} integrates agentic tool use with faithfulness evaluation, though it targets open-ended questions and relies on an LLM-as-a-judge pipeline. Beyond explainability, recent multimodal agentic systems have explored actor–critic architectures with tool grounding and iterative feedback for complex multi-step tasks \citep{ghosh2026carepilot,halder2026arogyasutra}. A parallel line applies multimodal agents to \emph{mechanistic} interpretability of internal model representations, including \textsc{Maia}~\citep{shaham2025multimodalautomatedinterpretabilityagent}, \textsc{OpenMaia}~\citep{camunas2025openmaia}, and \textsc{NeuronScope}~\citep{liu2026neuronscopemultiagentframeworkexplaining}; however, these are \textbf{not} designed for the prediction-level questions domain practitioners ask.

\noindent\underline{\smash{\textit{\textbf{Research Gap.}}}} No existing work jointly optimizes a multi-agent pipeline end-to-end with RL across diverse modalities and question types. Our framework addresses this via a Proposer–Actor structure trained with GRPO, using faithfulness as the reward signal.

%% file: 030method.tex
\section{Our Framework: \method}
\looseness=-1\xhdr{Problem Formulation} We study natural-language explanation generation for black-box classifiers. Given a trained classifier $f:\mathcal{X}\rightarrow\mathcal{Y}$, a question $q$ from a predefined taxonomy $\mathcal{Q}$, and input instance(s) $x \in \mathcal{X}$, \method generates a structured explanation $e$ that answers $q$ while remaining faithful to $f$'s decision process. \method supports three modalities (vision, text, and tabular data), where $x$ is an image, token sequence, or feature vector respectively, and localizes evidence accordingly: a bounding box, token spans, or feature keys, identifying the subset of $x$ that accounts for $f$'s behavior with respect to $q$.

\looseness=-1\xhdr{XAI Question Taxonomy}
\label{sec:question_taxonomy}
Drawing on prior work~\citep{10.1145/3734191,li2023mathcalm}, we define ten question types in three semantic categories (Appendix Table~\ref{tab:questions_template}): 1) \textit{Feature Attribution} asks which input parts drove or suppressed a prediction; 2) \textit{Counterfactual} probes prediction sensitivity to hypothetical modifications; and 3) \textit{Spurious Feature} examines whether the model relies on irrelevant cues. We exclude Q8 and Q9 from training and use them as \emph{out-of-distribution} (OOD) evaluation questions, since these are only defined over \emph{misclassified} instances ($\hat{y} \neq y^*$), which limits eligible training samples and destabilizes gradient estimation; we treat them as a held-out probe of generalization to unseen question types without task-specific supervision.


\subsection{\method Overview}
\looseness=-1\method operates as a two-stage framework on a shared multimodal LLM backbone~$\theta$. First, the \textit{Proposer Agent}~$\pi^{\mathrm{Prop}}_{\theta}$ receives question~$q$, input~$\mathbf{x}$, and prediction~$\hat{y} = f(\mathbf{x})$, and produces a structured strategy~$S$ specifying which XAI tools and autonomous reasoning tasks to invoke (Appendix~\ref{sec:xai_tools}). Then, the \textit{Actor Agent}~$\pi^{\mathrm{Act}}_{\theta}$ executes $S$, processes the raw tool outputs, and synthesizes an explanation~$e$:
\begin{align}
    S &\;\sim\; \pi^{\mathrm{Prop}}_{\theta}
               \!\left(S \;\middle|\; q,\,\mathbf{x},\,\hat{y},\,\mathcal{T}_{m}\right),
               \label{eq:proposer} \\[4pt]
    e &\;\sim\; \pi^{\mathrm{Act}}_{\theta}
               \!\left(e \;\middle|\; q,\,\mathbf{x},\,\hat{y},\,S,\,
               \{o_t\}_{t \in \mathcal{T}_{\mathrm{sel}}}\right),
               \label{eq:actor}
\end{align}
where $\mathcal{T}_{m}$ is the tool set for modality~$m$, $\mathcal{T}_{\mathrm{sel}} \subseteq \mathcal{T}_{m}$ is the Proposer's selected subset, and $o_t$ is tool~$t$'s summarized output.

\subsection{Proposer Agent: Intent-Aware Strategy Formation}
\looseness=-1 Prior conversational systems rely on fixed pipelines or hand-crafted routing rules with no learned component adapting to question semantics. To address this limitation, we introduce the trainable {Proposer Agent}~$\pi^{\mathrm{Prop}}_{\theta}$, which performs strategy formation before any tool is executed. Given question type~$Q_i$, modality~$m$, and prediction~$\hat{y} = f(\mathbf{x})$, it generates $S = (s_{\mathrm{type}},\,\mathcal{T}_{\mathrm{sel}},\,
\mathcal{A}_{\mathrm{auto}})$:
\begin{equation}
    S \;\sim\; \pi^{\mathrm{Prop}}_{\theta}
    \!\left(S \;\middle|\; q,\,\mathbf{x},\,\hat{y},\,\mathcal{T}_{m}\right),
    \label{eq:proposer}
\end{equation}
where $s_{\mathrm{type}} \in \{\texttt{autonomous},\,\texttt{tools},\,\texttt{hybrid}\}$ selects the reasoning mode, $\mathcal{T}_{\mathrm{sel}} \subseteq \mathcal{T}_{m}$ specifies which tools to invoke, and $\mathcal{A}_{\mathrm{auto}}$ enumerates grounding/reasoning subtasks for the Actor. An ablation on the Proposer's strategy modes (Appendix~\ref{sec:proposer_mode_ablation}) confirms both tool use and autonomous reasoning are independently load-bearing, and that \method learns to combine them more effectively than the base agent. \par

\subsection{Actor Agent: Tool-Grounded Explanation Synthesis}
\label{sec:actor_agent}
\looseness=-1 Even when a well-formed strategy is available, raw tool outputs are not directly consumable as explanations. \citet{kroeger2024incontextexplainersharnessingllms} show that LLMs systematically bypass tool evidence in favor of parametric knowledge, producing plausible but unfaithful conclusions. To address this, the {Actor Agent}~$\pi^{\mathrm{Act}}_{\theta}$ operates in two steps enforcing a structural dependency on attribution evidence.

\xhdr{Step 1: Tool Execution and Summarization} For each tool $t \in \mathcal{T}_{\mathrm{sel}}$, the Actor invokes $t$ against $f$ and $\mathbf{x}$, obtaining heatmaps, per-token, or feature importance scores, which are converted into structured text summaries (execution status, natural-language description, key statistics) rather than passed as raw tensors. To standardize granularity, the Actor returns the top~25\% of features/tokens for tabular/text inputs, and aggregates top-1\% attribution coordinates into a bounding box for vision.

\looseness=-1\xhdr{Step 2: Evidence-Grounded Explanation Generation} The per-tool summaries are assembled into a structured prompt over which the Actor reasons to identify the most relevant features:
\begin{equation}
    e \;\sim\; \pi^{\mathrm{Act}}_{\theta}
    \!\left(e \;\middle|\; q,\,\mathbf{x},\,\hat{y},\,S,\,
    \{o_t\}_{t \in \mathcal{T}_{\mathrm{sel}}}\right), \quad e \in \mathcal{E}_{m},
    \label{eq:actor}
\end{equation}
where $\mathcal{E}_{m}$ is the modality-appropriate output space: a bounding box, token span, or feature list.\par 

\subsection{Perturbation-based Faithfulness Evaluation}
\label{sec:metrics}
Evaluating explanation correctness without ground-truth labels is difficult. Prior work relies on LLM-as-a-judge pipelines susceptible to the same hallucinations they assess~\citep{kroeger2024incontextexplainersharnessingllms, Rawal_2025}, or evaluates a single explanation type in isolation~\citep{li2023mathcalm,kim2025towards}; none adapts faithfulness evaluation into an online training signal across a multi-question, multi-modal setup. We address this with a unified {perturbation-based protocol}: explanation $e$ identifying region $r$ is faithful iff masking $r$ produces the model behavior change $q$ predicts. For each $Q_i$, metric $m_i$ is computed by masking $r$ in $\mathbf{x}$ and re-querying $f$, where $\mathbf{x}^{-r}$ is the masked input (zero-filled for tabular, grey-filled for vision, span-deleted for text) and $P(y \mid \mathbf{x})$ the predicted class probability.

\xhdr{Feature Attribution~(Q1-Q4)}
Q1-Q2 form a complementary pair: masking the most responsible region should drop confidence; masking the least responsible should not:
\begin{align}
    m_1 &=
    \max\!\left(
    0,\;
    P(\hat{y} \mid \mathbf{x})
    - P(\hat{y} \mid \mathbf{x}^{-r})
    \right), \nonumber \\
    m_2 &=
    1 -
    \left|
    P(\hat{y} \mid \mathbf{x})
    - P(\hat{y} \mid \mathbf{x}^{-r})
    \right|.
\end{align}

\looseness=-1 Q3 checks whether masking the distinctive region collapses the top-1/top-2 margin; Q4 measures cross-instance contrastiveness via $\mathrm{sim}(F_A, F_B)$, requiring different attributions for differently-predicted instances.

\looseness=-1\xhdr{Counterfactual~(Q5--Q7)} Q5 verifies the predicted direction of change agrees with the observed probability shift; Q6-Q7 measure whether the proposed modification raises the target-class probability and whether the predicted new class becomes likely:
\begin{align}
    m_6 &=
    \max\!\left(
    0,\;
    P(\hat{y} \mid \mathbf{x}^{+\Delta})
    - P(\hat{y} \mid \mathbf{x})
    \right), \nonumber \\
    m_7 &=
    P(\hat{y}' \mid \mathbf{x}^{-r}),
\end{align}
where $\mathbf{x}^{+\Delta}$ is the agent-proposed modification and $\hat{y}'$ the predicted class.

\looseness=-1\xhdr{Spurious Features} Q8 measures the normalized gain in correct-class probability after masking:
\begin{align}
    m_8 &=
    \max\!\left(
    0,\;
    \min\!\left(1,\; D(\mathbf{x}, r)\right)
    \right), \nonumber \\
    D(\mathbf{x}, r)
    &=
    \frac{
    P(y^{*} \mid \mathbf{x}^{-r})
    -
    P(y^{*} \mid \mathbf{x})
    }{
    1 - P(y^{*} \mid \mathbf{x})
    },
\end{align}
normalized by remaining headroom for fair cross-instance comparison. Q9 averages $m_8$ over $N$ misclassified instances; Q10 mirrors Q4's contrastive intuition on misclassified instances. Full definitions are in Appendix~\ref{sec:metrics_details}.

\subsection{End-to-End Training}
\looseness=-1 
Reward-based optimization for trustworthy generation has increasingly relied on carefully designed task-specific objectives \citep{ghosh2026rado}. However relying exclusively on raw faithfulness scores frequently precipitates reward hacking, typically via two degenerate strategies: trivial probability-drop maximization (\eg masking an entire image) and indiscriminate tool invocation, which inflates faithfulness without isolating informative features~\citep{shao2024deepseekmathpushinglimitsmathematical}. Since naive objectives converge on these unfaithful traces, we employ GRPO~\citep{guo2025deepseek} to jointly optimize both agents end-to-end , building on recent applications of GRPO for reasoning optimization\citep{onyame2026cure}. Each episode captures two transitions: strategy generation by the Proposer and explanation synthesis by the Actor. We formalize this as an MDP with deterministic tool-execution transitions and a terminal reward in Appendix~\ref{sec:mdp_formalization}. The faithfulness reward~$m_i$ is evaluated strictly after the Actor's final output, and the trajectory-level group-normalized advantage
\begin{equation}
    \hat{A}_k = \frac{R_k - \mathrm{mean}\left(\{R_j\}_{j=1}^{G}\right)}
                     {\mathrm{std}\left(\{R_j\}_{j=1}^{G}\right)},
    \label{eq:advantage}
\end{equation}
is applied uniformly to both transitions, treating the Proposer-Actor as jointly responsible for the outcome: a misaligned Proposer strategy cannot be recovered by the Actor, so the optimization signal must flow to both agents equally.

\looseness=-1\xhdr{Size Penalty (Vision)} To foreclose the large-mask shortcut, we penalize large bounding boxes:
\begin{equation}
    m_i^{\mathrm{vision}} = \max\!\left(0,\; m_i - \lambda_{\mathrm{size}} \cdot \rho\right),
    \label{eq:size_penalty}
\end{equation}
where $\rho \in [0,1]$ is the bounding-box-to-image area ratio and $\lambda_{\mathrm{size}} = 0.3$. For tabular/text, the fixed top-25\% output constraint~(Section~\ref{sec:actor_agent}) already governs scope, making an explicit size penalty unnecessary.

\looseness=-1\xhdr{Tool Penalty} We apply an $\ell_1$ tool-count penalty across all modalities: $r{=}\max\!\left(0,\;\min\!\left(1,\; m_i{-}\lambda_{\mathrm{tool}} \cdot \frac{n_{\mathrm{tools}}}{n_{\mathrm{max}}}\right)\right)$, where $n_{\mathrm{tools}}$ is the number of invoked tools/tasks, $n_{\mathrm{max}}{=}9$, $\lambda_{\mathrm{tool}}{=}0.05$--kept small to avoid suppressing gradients on question types with low pre-training scores (\eg tabular Q1/Q8: 0.01/0.04 at initialization). These penalties let the Proposer learn informative tool selection while the Actor learns faithful grounding, each reinforcing the other via shared gradients.

%% file: 040results.tex
\section{Experiments}
\looseness=-1 We address the following key research questions: {RQ1)} Does \method generates faithful explanation across all three modalities? {RQ2)} Does \method outperform widely used post hoc explainers? {RQ3)} Is the prompting style and agentic architecture necessary for \method's performance? and {RQ4)} Does \method generalize to OOD explanation questions, unseen datasets, and novel model architectures?

\subsection{Datasets and Experimental Setup}
\begin{figure*}
    \centering
    \includegraphics[width=\linewidth]{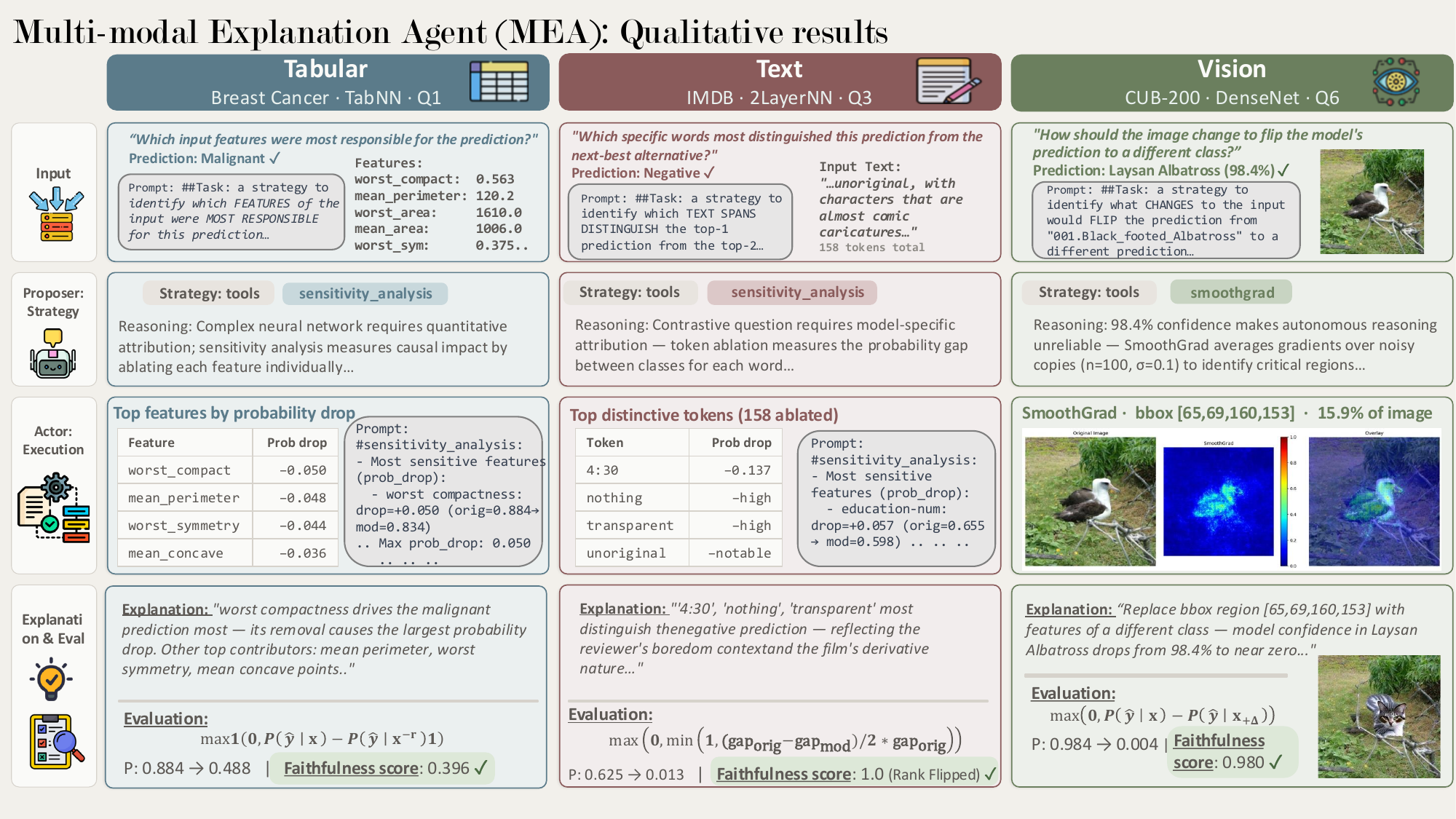}
    \vspace{-0.15in}
    \caption{\looseness=-1\xhdr{Case study of \method across three modalities} Each column shows a complete pipeline execution for \method: the input, Proposer's strategy and tool, Actor's execution output, and final explanation with faithfulness evaluation.}
    \vspace{-0.075in}
    \label{fig:case_study}
\end{figure*}

We first describe datasets designed to study the faithfulness of explanations generated by \method and then outline experimental setup. Figure~\ref{fig:case_study} illustrates \method's end-to-end pipeline on representative samples from each modality, showing how the Proposer selects appropriate tools and the Actor synthesizes faithful explanations across diverse question types.

\looseness=-1\xhdr{Dataset} We construct a multi-modal benchmark for analyzing agent behavior across tabular, text, and vision modalities (Appendix~\ref{sec:dataset_app}), generated using predictions from ML models trained on standard supervised learning tasks. For each dataset-model pair, we generate structured questions (Table~\ref{tab:questions_template}) covering feature attribution, counterfactual analysis, and spurious feature identification. \textbf{Tabular}: Adult Census Income~\citep{adultcensus} and Breast Cancer~\citep{breastcancer}, evaluated with a two-layer network and TabNet. \textbf{Text}: SNLI~\citep{bowman2015largeannotatedcorpuslearning} and IMDb~\citep{imdb}, evaluated with a lightweight neural network and a CNN. \textbf{Vision}: CUB-200-2011~\citep{welinder2010cub200} and STL-10~\citep{stl10}, evaluated with ResNet-50~\citep{resnet} and DenseNet-201~\citep{densenet}. Each instance contains the original input, ground-truth label, model prediction, and a natural-language analytical prompt.

\looseness=-1\xhdr{Training Setup} We fine-tune Qwen3.6-35B-A3B~\citep{qwen36_35b_a3b} using LoRA ($r{=}32$)~\citep{hu2021loralowrankadaptationlarge} for 3 epochs, following \citet{shao2024deepseekmathpushinglimitsmathematical} for GRPO hyperparameters (learning rate $1\times 10^{-5}$, batch size 32, KL coefficient 0). We generate $K{=}4$ rollouts per question at $T{=}1.0$ ($K{=}8$ for harder types), with advantages group-normalized over each question's $K$ rollouts; evaluation uses $T{=}0.0$. For training across modalities, we compare \textit{Sequential} (tabular$\to$text$\to$vision, each stage initialized from the previous checkpoint) against \textit{Hybrid} (all modalities interleaved per batch). Sequential training suffers from catastrophic forgetting. Performance on earlier modalities degrades and behavioral diversity collapses on the final modality (Appendix~\ref{sec:training_strategy}), so we adopt Hybrid training for all main results. We also ablate on a 4B-parameter backbone (Appendix~\ref{sec:backbone_scale}), confirming faithfulness gains transfer to smaller models; computational cost details are in Appendix~\ref{sec:app_computation_cost}.

\looseness=-1\xhdr{Baselines} To isolate the impact of collaborative, iterative refinement, we benchmark \method against four prompting paradigms sharing identical model infrastructure, execution environments, and XAI toolsets: \emph{Zero-Shot} (control), \emph{Chain-of-Thought (CoT)}~\citep{wei2023chainofthoughtpromptingelicitsreasoning}, \emph{ReAct}~\citep{yao2023reactsynergizingreasoningacting}, and \emph{Tree of Thoughts (ToT)}~\citep{yao2023treethoughtsdeliberateproblem}. We further compare against standard post-hoc explainers~\citep{ribeiro2016whyitrustyou,lundberg2017unifiedapproachinterpretingmodel, smilkov2017smoothgradremovingnoiseadding, Selvaraju_2019, springenberg2015strivingsimplicityconvolutionalnet, article, sundararajan2017axiomaticattributiondeepnetworks}, and against specialized counterfactual generators, DiCE~\citep{mothilal2020explaining} and Growing Spheres~\citep{laugel2017inverse}.

\begin{figure*}[tp]
\centering 
\includegraphics[width=\linewidth]{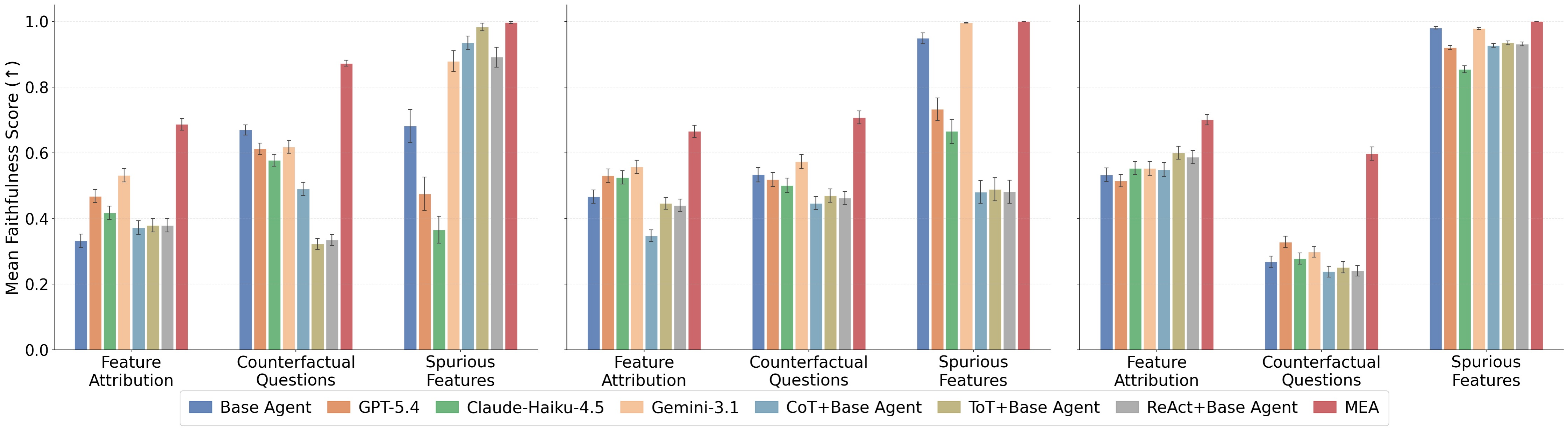}
\vspace{-0.1in}
\caption{\looseness=-1\xhdr{\method's performance} Faithfulness scores of \method vs. baselines across three modalities: \textbf{(A)} tabular, \textbf{(B)} text, and \textbf{(C)} vision. Higher mean faithfulness scores indicate more faithful explanations. 
Results show that \method outperform all baselines, including closed-source models and techniques like CoT, ToT, and ReACT.}
\vspace{-0.1in}
\label{fig:results_main}
\end{figure*}

\looseness=-1\xhdr{Evaluation} Following~\citet{agarwal2022openxai}, we evaluate explanation quality through \emph{faithfulness}: whether the agent's claimed explanation is verified by the target model's behavior under perturbation. All metrics (Sec.~\ref{sec:metrics}) follow a common \emph{perturb-and-measure} protocol: given an explanation identifying a region $r$ (feature keys for tabular, text spans for text, a bounding box for vision), we perturb $r$ to obtain $x'$, re-run the prediction, and score it against the corresponding metric. Perturbations are modality-specific: vision regions are gray-filled (or inpainted with Stable Diffusion~\citep{rombach2022highresolutionimagesynthesislatent} for counterfactuals), text spans are deleted or replaced with agent-provided counterfactual text, and tabular features are mean-filled or set to agent-specified values.

\section{Results}
\label{sec:results}

Here, we discuss experimental results that answer key questions (RQ1-RQ4) highlighted before.

\xhdr{RQ1) \method improves explanation faithfulness across all modalities} A key challenge in XAI is that different explanation needs (feature attribution, counterfactual reasoning, and spurious feature detection) traditionally require separate tools. \method addresses this by serving as a unified framework handling diverse question types across all three modalities without modality-specific engineering. Fig.~\ref{fig:results_main} shows \method achieves the highest faithfulness scores across all three modalities. On tabular data, \method scores $0.855$ on counterfactual questions, outperforming the best baseline (Base Agent, $0.670$) by \textbf{28\%}, and $0.997$ on spurious features vs.\ Gemini-3.1's $0.879$. On text, \method scores $0.708$ on counterfactual questions vs.\ Gemini-3.1's $0.574$ (\textbf{+23\%}). The gap is largest on vision counterfactual questions, where \method achieves $0.598$ vs.\ GPT-5.4's $0.328$ (\textbf{+82\%}). Across all three modalities, \method attains a perfect or near-perfect spurious feature detection score ($0.997$/$1.000$/$1.000$). To verify these similarities reflect genuine statistical equivalence rather than noise, we ran two-sided Welch's $t$-tests~\citep{10.1093/biomet/34.1-2.28} on all targeted comparisons: on spurious-feature questions (Q10), \method achieves ceiling performance ($1.000$) on text and vision, and the strongest baselines (Gemini-3.1: $0.996$ text, Base Agent: $0.981$ vision) differ by at most $0.02$--significant only due to \method's zero variance at ceiling (Appendix~\ref{sec:app_significance}). Appendix~\ref{sec:bias_sanity_check} further shows this faithfulness tracks the target model's decision logic, not the LLM's own priors.

These results confirm that \method acts as a \texttt{one-stop-shop} delivering strong, reliable faithfulness across all ten question types and all three modalities. \method also outperforms all closed-source models and alternative agentic frameworks (CoT, ToT, ReAct) on all question types across all modalities, confirming its role as a unified solution for diverse explainability needs.

\begin{figure}[tp]
\centering
\includegraphics[width=0.9\linewidth]{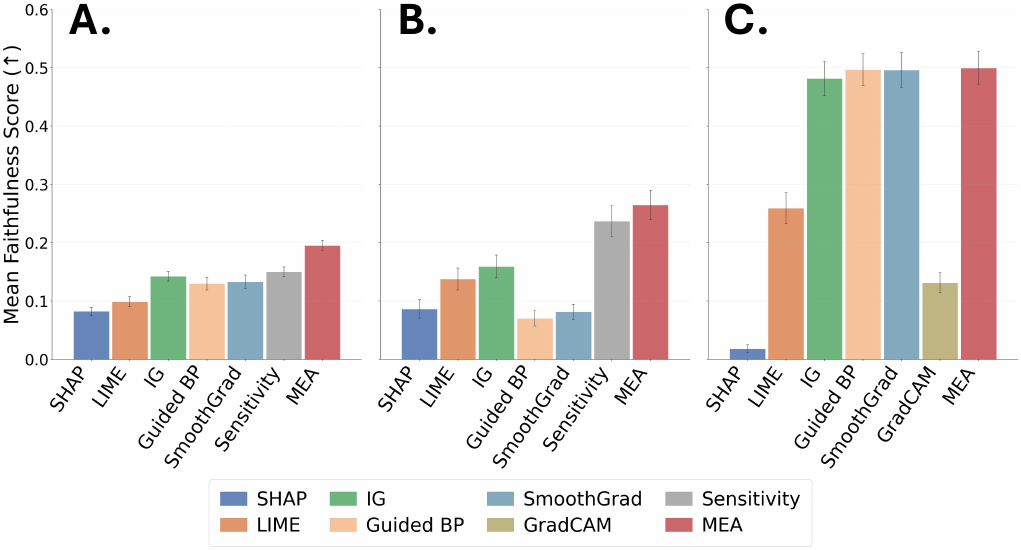}
\vspace{-0.1in}
\caption{\looseness=-1\xhdr{\method vs. Explainers} Mean faithfulness scores of identifying most salient input features across \textbf{(A)} tabular, \textbf{(B)} text, and \textbf{(C)} vision modalities. \method outperforms all baselines on tabular and text inputs, and matches the best-performing method on vision inputs.
}\vspace{-0.1in}
\label{fig:tool_only_baseline}
\end{figure}

\looseness=-1\xhdr{RQ2) \method outperforms post-hoc explainers}
Standard post-hoc explainers (LIME, SHAP, GradCAM, etc.) are architecturally single-purpose, so a like-for-like comparison is well-defined only for Q1: \textit{Which parts of the instance are most responsible for the model's prediction?} This restriction itself illustrates one of \method's motivations: no single tool can address the diverse explanation needs spanned by \method's ten-question taxonomy. We therefore compare \method against two \emph{disjoint} families of specialized baselines, each matched to the question type it targets: general-purpose attribution explainers on Q1, and specialized counterfactual generators on Q6.

\textbf{Q1 (Feature Attribution).} Following the same output standardization used in \method (top-25\% of features for tabular/text, top-1\% pixels as a bounding box for vision), Fig.~\ref{fig:tool_only_baseline} shows \method outperforms the best post-hoc explainer on tabular ($0.195$ vs.\ $0.150$; \textbf{+30\%}) and text ($0.265$ vs.\ $0.237$; \textbf{+12\%}), and matches the best single-tool baseline on vision ($0.499$ vs.\ $0.497$; $p=0.944$, Appendix~\ref{sec:app_significance}).

\textbf{Q6 (Flip Prediction).} We further compare \method against two specialized counterfactual generators, DiCE~\citep{mothilal2020explaining} and Growing Spheres~\citep{laugel2017inverse}. Neither has a meaningful notion of searching over text spans or image regions, so this comparison is restricted to \emph{tabular} data, using the same test instances, preprocessor, and metrics as \method's evaluation (Table~\ref{tab:dice_comparison}). \method achieves comparable flip-success rates while achieving substantially higher faithfulness throughout (avg.\ target-class probability $0.886$--$0.894$ vs.\ $0.499$--$0.631$).

\begin{table}[tp]
\centering
\small
\caption{\method vs.\ specialized counterfactual explainers on Q6 (flip prediction), aggregated across model architectures per dataset ($n=63$ for Adult Census, $n=44$ for Breast Cancer). Best per row group in \textbf{bold}.}
\label{tab:dice_comparison}
\begin{tabular}{llcc}
\toprule
\textbf{Dataset} & \textbf{Method} & \textbf{Flip Success} & \textbf{Avg.\ Faithfulness} \\
\midrule
\multirow{3}{*}{Adult Census} & \method & 100.0\% & \textbf{0.894} \\
& DiCE & 100.0\% & 0.631 \\
& Growing Spheres & 98.4\% & 0.499 \\
\midrule
\multirow{3}{*}{Breast Cancer} & \method & 95.5\% & \textbf{0.886} \\
& DiCE & 97.7\% & 0.580 \\
& Growing Spheres & 81.8\% & 0.412 \\
\bottomrule
\end{tabular}
\end{table}

Together, these results show that \method, despite operating as a single \textbf{general-purpose agent} across all ten question types, is competitive with or superior to specialized single-purpose explainers on both feature attribution and counterfactual generation.

\looseness=-1\xhdr{RQ3) \method is robust to different prompt styles and outperforms critic-augmented agents} Here, we first test sensitivity to question phrasing by evaluating three paraphrased variants of the question taxonomy (see \ref{sec:paraphrase}) and compare mean faithfulness against the original. Next, we investigate whether augmenting a base agent with a Critic~\citep{shinn2023reflexionlanguageagentsverbal}, which receives the original strategy and explanation, critiques them, and regenerates improved outputs, can close the gap with \method. Results in Appendix ~\ref{sec:paraphrase} Fig.~\ref{fig:ablation_paraphrase} show that paraphrase variants score within $\mathbf{<1\%}$ of the original across all modalities and question groups (\eg tabular counterfactual: paraphrase $0.934$ vs.\ \method $0.939$; vision feature attribution: $0.686$ vs.\ $0.692$), confirming that \method is robust to prompt rephrasing. \method also consistently outperforms a base agent augmented with an inference-time Critic (Agent+Reflection) across all three modalities, with the largest gains on counterfactual questions: \textbf{+36\%} on text ($0.876$ vs.\ $0.644$) and \textbf{+30\%} on vision ($0.569$ vs.\ $0.439$), and Agent+Reflection even \emph{degrading} performance relative to the untrained base agent on tabular spurious features (Appendix~\ref{sec:critic_agent}, Fig.~\ref{fig:ablation_reflection}). Appendix ~\ref{sec:app_significance} shows paraphrase robustness tests results, which demonstrate no significant performance degradation
across 8 of 9 question-type$\times$modality conditions ($p>0.05$ in all but vision counterfactual). These results demonstrate that RL-based end-to-end training provides substantially stronger and more reliable gains than inference-time critique.

\looseness=-1\xhdr{RQ4) \method generalizes to OOD datasets and model architectures} We evaluate \method on unseen datasets and model architectures
across all modalities (Yelp Review Polarity dataset~\cite{zhang2016characterlevelconvolutionalnetworkstext} with a BERT-based classifier~\cite{devlin2019bertpretrainingdeepbidirectional} for text; German Credit dataset~\cite{dua2019uci} with a three-layer neural network for tabular; CIFAR-10~\cite{krizhevsky2009learning} with ResNet-18~\cite{resnet} for vision). Results in Table~\ref{tab:ood_data_faithfulness} show that \method achieves the highest overall faithfulness on all three modalities ($0.581$ vs. $0.379$ on tabular, $0.689$ vs. $0.625$ on text, $0.576$ vs. $0.448$ on vision). The largest gains appear on tabular inputs ($+\textbf{53\%}$), with feature attribution alone rising from $0.447$ to $0.783$ (\textbf{+75\%}). On vision, counterfactual faithfulness improves from $0.246$ to $0.521$ (\textbf{+112\%}). \method only underperforms the base agent on text counterfactual questions ($0.770$ vs. $0.860$) and vision feature attribution ($0.583$ vs.\ $0.621$), which we attribute to further hyperparameter tuning needed for better modality-specific trade-offs. \method also generalizes to concept-level explanation, a question type outside the trained taxonomy evaluated via representation-space CAV ablation, outperforming the base agent by $\sim$18\% in mean faithfulness. Overall, these results confirm that \method \textbf{generalizes} to unseen datasets and model architectures, and to explanation mechanisms beyond input perturbation (see Appendix Table~\ref{app:ood_data_faithfulness} for detailed results, Table~\ref{tab:q11_concept} for concept-level results, and Appendix \ref{sec:ood_ablation} for an ablation study of OOD question types).

\begin{table}[tp]
\centering
\small
\caption{\looseness=-1 OOD faithfulness on datasets/architectures unseen during training, averaged across question types per group. \textbf{Base}: Proposer--Actor without GRPO fine-tuning; \textbf{\method}: full GRPO-tuned system. Best per row in \textbf{bold}.}
\label{tab:ood_data_faithfulness}
\small
\setlength{\tabcolsep}{4pt}
\begin{tabular}{llcc}
\toprule
\textbf{Modality} & \textbf{Group} & \textbf{Base} & \textbf{\method} \\
\midrule
\multirow{4}{*}{Tabular} & Feat. Attr. & 0.447 & \textbf{0.783} \\
& Counterfactual & 0.422 & \textbf{0.499} \\
& Spurious Feat. & 0.264 & \textbf{0.349} \\
\cline{2-4}
& \textit{Overall} & 0.379 & \textbf{0.581} \\
\midrule
\multirow{4}{*}{Text} & Feat. Attr. & 0.428 & \textbf{0.563} \\
& Counterfactual & \textbf{0.860} & 0.770 \\
& Spurious Feat. & 0.639 & \textbf{0.778} \\
\cline{2-4}
& \textit{Overall} & 0.625 & \textbf{0.689} \\
\midrule
\multirow{4}{*}{Vision} & Feat. Attr. & \textbf{0.621} & 0.583 \\
& Counterfactual & 0.246 & \textbf{0.521} \\
& Spurious Feat. & 0.416 & \textbf{0.624} \\
\cline{2-4}
& \textit{Overall} & 0.448 & \textbf{0.576} \\
\bottomrule
\end{tabular}
\end{table}

%% file: 050conclusion.tex
\section{Conclusion}
\label{sec:conclusion}
\looseness=-1 We presented \method, a multi-agent framework for faithful, prediction-level explanation across tabular, text, and vision modalities, and \methodbench, a model- and modality-agnostic benchmark of 16k instances each paired with a faithfulness metric. Optimizing \method end-to-end via RL with a modality-adaptive reward yields an agent that consistently outperforms post-hoc explainers, frontier models, and advanced agentic frameworks by $+28\%$ (tabular), $+21\%$ (text), and $+34\%$ (vision) faithfulness, showing \textbf{learning to explain} is more principled than prompting to explain. \method is currently scoped to classification tasks and a fixed tool library, with a perturbation-based protocol sensitive to distributional shift like others in its class; extending to other tasks, tool libraries, and faithfulness signals is promising future work.

%% file: 060appendix.tex
\appendix
\section{Appendix}
The Appendix provides supplementary material supporting the main paper, including the complete question taxonomy (\ref{sec:appendix_taxonomy}), details of the XAI tools employed (\ref{sec:xai_tools}), faithfulness evaluation metrics (\ref{sec:metrics_details}), a formal RL problem definition (\ref{sec:mdp_formalization}), dataset statistics (\ref{sec:dataset_app}), descriptions of the sequential and hybrid training procedures (\ref{sec:training_strategy}), ablation studies on out-of-distribution (OOD) question types, unseen datasets/architectures, and concept-level explanation generalization (\ref{sec:ood_ablation}), prompt robustness (\ref{sec:paraphrase}), the critic agent (\ref{sec:critic_agent}), the Proposer's strategy modes (\ref{sec:proposer_mode_ablation}), backbone scale transfer (\ref{sec:backbone_scale}), a sanity check on a classifier with a synthetic shortcut (\ref{sec:bias_sanity_check}), statistical significance analyses (\ref{sec:app_significance}), computational cost details (\ref{sec:app_computation_cost}) and prompt templates for Actor and Proposer Agents (\ref{sec:prompts}).

\subsection{Question Taxonomy}
The detailed question definition and category is showed in Table \ref{tab:questions_template}.
\label{sec:appendix_taxonomy}
\begin{table*}[t]
\centering
\small
\renewcommand{\arraystretch}{1.3}
\caption{Overview of the ten explanation questions organised by category.}
\label{tab:questions_template}
\begin{tabular}{p{0.6cm} p{2.8cm} p{9.8cm}}
\hline
\textbf{ID} & \textbf{Category} & \textbf{Question} \\
\hline
\multicolumn{3}{l}{\textit{Feature Attribution - which parts of the input drove or suppressed the prediction?}} \\
\hline
Q1 & Most Responsible   & Which parts of the instance are most responsible for the model's prediction? \\
Q2 & Least Responsible  & Which parts of the instance are least responsible for the model's prediction? \\
Q3 & Distinctive        & Which specific parts of the instance distinguish its prediction from the next-best alternative? \\
Q4 & Contrastive        & Why are instances $A$ and $B$ given different predictions? \\
\hline
\multicolumn{3}{l}{\textit{Counterfactual - how sensitive is the prediction to hypothetical modifications?}} \\
\hline
Q5 & Mask Prediction    & If we mask a certain part of this instance, would the prediction change? \\
Q6 & Flip Prediction    & How should the instance change to flip the model into a different prediction? \\
Q7 & Change Prediction  & If we remove or change one important part of the instance, how would the prediction change? \\
\hline
\multicolumn{3}{l}{\textit{Spurious Feature - does the model rely on irrelevant cues?}} \\
\hline
Q8  & Irrelevant Parts   & Is there any irrelevant part in this instance that causes the model's wrong prediction? \\
Q9  & Shared Feature     & What shared feature makes misclassified instances $A$, $B$, $C$, \ldots difficult for the model? \\
Q10 & Similar-Different  & Why are instances $A$ and $B$ assigned different predictions despite apparent similarity? \\
\hline
\end{tabular}
\end{table*}

\subsection{XAI Tools}
\label{sec:xai_tools}
To support explanation generation, our framework provides a library of established attribution methods spanning all three modalities:

\xhdr{Gradient-based} GradCAM \cite{Selvaraju_2019} (layer-wise gradient attribution, vision only), Integrated Gradients \cite{sundararajan2017axiomaticattributiondeepnetworks}, Guided Backpropagation \cite{springenberg2015strivingsimplicityconvolutionalnet}, SmoothGrad \cite{smilkov2017smoothgradremovingnoiseadding} (noise-averaged gradient estimation), and Sensitivity Analysis \cite{article} (input-gradient saliency, text and tabular only).

\xhdr{Perturbation-based} LIME \cite{ribeiro2016whyitrustyou} (Local Interpretable Model-agnostic Explanations with superpixel or token-level segmentation for vision, and token-level perturbation for text) and SHAP \cite{lundberg2017unifiedapproachinterpretingmodel} (Shapley Additive exPlanations via marginal contribution estimation).

\subsection{Faithfulness Metrics}
\label{sec:metrics_details}
\begin{table*}[t]
\centering
\small
\caption{
Summary of evaluation metrics corresponding to different explanation faithfulness questions.
}
\label{tab:faithfulness_metrics}
\begin{tabular}{p{1.2cm} p{6cm} p{5.2cm}}
\hline
\textbf{Question} & \textbf{Metric $m_i$} & \textbf{Intuition} \\
\hline
Q1 &
$\max\bigl(0,P(\hat{y}\mid\mathbf{x}) -P(\hat{y}\mid\mathbf{x}^{-r})\bigr)$
&
Masking the most responsible region should drop predicted-class confidence
\\
Q2 &
$1 - \bigl\lvert P(\hat{y}\mid\mathbf{x}) - P(\hat{y}\mid\mathbf{x}^{-r})\bigr\rvert$
&
Masking the least responsible region should leave confidence unchanged
\\
Q3 & $\max\!\left(0,\min\!\left(1,\dfrac{\text{gap}_{\text{orig}}-\text{gap}_{\text{mod}}}{2*\text{gap}_{\text{orig}}}\right)\right)$
&
The top-1/top-2 margin should collapse after masking the distinctive region
\\
Q4 &
$1 - \mathrm{sim}(F_A, F_B)$
&
The agent should attribute different features to each instance
\\
Q5 &
$\begin{cases} \max(0,\;\delta) & \hat{c}=1\ \text{(predicts change)} \\
\max(0,\;1-\lvert\delta\rvert) & \hat{c}=0\ \text{(predicts stable)} \end{cases} $
&
Probability change should agree in direction with the agent's yes/no answer
\\
Q6 &
$\begin{cases} \max(0,P(\hat{y}\mid\mathbf{x}) - P(\hat{y}\mid\mathbf{x}_{+\Delta})) & \text{vision} \\ \max(0,\min(1,P(y_t\mid\mathbf{x}_{+\Delta}))) & \text{text/tab} \end{cases}$
&
Applying the proposed change should raise the probability of the target class
\\
Q7 &
$P(\hat{y} \mid \mathbf{x}^{-r})$
&
The agent's predicted new class should have high probability after masking
\\
Q8 &
$\max\!\left(0,\min\!\left(1,\dfrac{P(y^*\mid\mathbf{x}^{-r})-P(y^*\mid\mathbf{x})}{1-P(y^*\mid\mathbf{x})}\right)\right)$
&
Masking the spurious region should improve correct-class probability
\\
Q9 &
$\dfrac{1}{N}\displaystyle\sum_{i=1}^{N}\max\!\left(0,\min\!\left(1,\dfrac{\Delta_i}{1-P(y^*_i\mid\mathbf{x}_i)}\right)\right)$
&
Improvement should hold across all $N$ misclassified instances
\\
Q10 &
$1 - \mathrm{sim}(F_{\text{corr}}, F_{\text{wrong}})$
&
Features attributed to the correct vs. wrong instance should differ
\\
\hline
\end{tabular}
\end{table*}

Here $\text{gap}_\text{orig}, \text{gap}_\text{mod}$ are the original gap between top-1 and top-2 predicted classes and the gap after perturbation; $y^*$ is the ground-truth class; $\hat{y}$ is the predicted class after perturbation; $\delta$ is the prediction probability drop after perturbation; and $F_{\text{correct}}, F_{\text{wrong}}$ are attribution maps for the correctly and incorrectly classified instances respectively. 

A higher metric score indicates a more faithful explanation. For binary metrics (Q3, Q5, Q6, Q7), the score is 1 if the condition holds and 0 otherwise. For Q6, tabular and text inputs are evaluated against a specific target class $y_t$, whereas vision inputs are evaluated by whether the prediction changes to any alternative class, reflecting the substantially larger label space in vision tasks.

The detailed metric calculation method and intuition is showed in Table \ref{tab:faithfulness_metrics}.

\subsection{Formal RL Problem Definition}
\label{sec:mdp_formalization}
We formalize \method's training as a two-step Markov Decision Process (MDP) over a single episode, with the Proposer and Actor as two sequential decision points sharing backbone $\theta$.

\xhdr{State (Proposer step)} $s^{\mathrm{Prop}} = (q,\,\mathbf{x},\,\hat{y},\,\mathcal{T}_{m})$: the question, input instance, frozen-classifier prediction, and the modality-compatible tool set (Sec.~\ref{sec:actor_agent}, Eq.~\ref{eq:proposer}).

\xhdr{Action (Proposer)} The structured strategy $S = (s_{\mathrm{type}},\,\mathcal{T}_{\mathrm{sel}},\,\mathcal{A}_{\mathrm{auto}})$ (Sec.~\ref{sec:actor_agent}). $\mathcal{T}_{\mathrm{sel}}$ and $\mathcal{A}_{\mathrm{auto}}$ are combinatorial subsets rather than a single discrete choice, so this is regarded as a structured action space.

\xhdr{Transition} Deterministic: each selected tool $t \in \mathcal{T}_{\mathrm{sel}}$ is executed against the frozen classifier $f$ and input $\mathbf{x}$, producing summarized outputs $\{o_t\}_{t \in \mathcal{T}_{\mathrm{sel}}}$ (Sec.~\ref{sec:actor_agent}, Step 1). No stochasticity is introduced by the environment; the only randomness in the episode comes from the two agents' own sampling.

\xhdr{State (Actor step)} $s^{\mathrm{Act}} = (q,\,\mathbf{x},\,\hat{y},\,S,\,\{o_t\}_{t \in \mathcal{T}_{\mathrm{sel}}})$: the Proposer's emitted strategy together with the summarized tool outputs produced by the transition above.

\xhdr{Action (Actor)} The explanation $e \in \mathcal{E}_{m}$ (Eq.~\ref{eq:actor}).

\xhdr{Reward} Terminal only: $R = m_i(e)$, the faithfulness metric for question type $Q_i$ (Sec.~\ref{sec:metrics}), augmented by the modality-adaptive size/tool penalties (Eq.~\ref{eq:size_penalty} and the tool penalty in Sec.~\ref{sec:actor_agent}), evaluated strictly after the Actor's final output. No intermediate reward is given at the Proposer's transition.

\xhdr{Shared parameterization} Both $\pi^{\mathrm{Prop}}_{\theta}$ and $\pi^{\mathrm{Act}}_{\theta}$ route through the \emph{same} LoRA-adapted backbone $\theta$; the two roles are differentiated entirely by distinct system prompts (Appendix~\ref{sec:prompts}, Figs.~\ref{fig:proposer_prompt}, \ref{fig:actor_prompt}) rather than separate adapters or a mode flag. Because $R$ is terminal and both transitions belong to the same episode, the group-normalized advantage $\hat{A}_k$ (Eq.~\ref{eq:advantage}) is applied identically to every token in both the Proposer's and the Actor's transitions, implementing the joint credit assignment described in Sec.~\ref{sec:actor_agent}.

\subsection{Dataset Statistics}
\label{sec:dataset_app}
 \begin{figure}[htbp]
    \centering
    \includegraphics[width=0.48\textwidth]{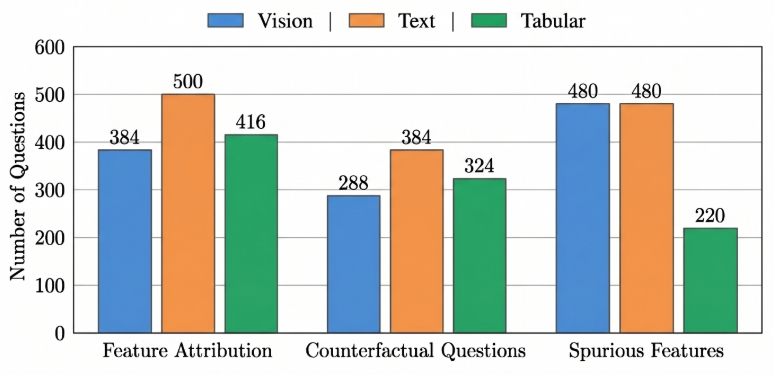}
    \caption{Dsitribution of data across modalities.}
    \label{fig:my_label}
\end{figure}

Due to the heterogeneous prediction behavior exhibited across models and datasets, the number of generated instances varies across categories (\textbf{Q1-Q10}). Different question types require specific prediction configurations, such as correct predictions, informative misclassifications, contrastive prediction pairs, or shared behavioral patterns across models. Consequently, the final instance count for each category is determined by the availability of meaningful and semantically valid examples satisfying these construction criteria. This naturally results in varying sample counts across question types, while ensuring that each generated instance reflects authentic model behavior and preserves the fidelity of the underlying prediction distributions.

\subsection{Sequential and Hybrid Training}
\label{sec:training_strategy}

We compare two multi-modal GRPO training schedules, both using Qwen3-VL-30B-A3B-Instruct\cite{Qwen-VL}.
\textit{Sequential} training fine-tunes one modality at a time, carrying each checkpoint forward to the next phase. \textit{Hybrid} training interleaves all three modalities within every batch, grouped by question type via Q-type stratification.

\subsection*{Sequential Training}

\paragraph{Phase 1 - Tabular.}
Fine-tuning on tabular data across 3 epochs drives the mean batch reward from ~0.3 to ~0.6. The resulting checkpoint achieves a tabular faithfulness score of 0.632.

\paragraph{Phase 2 - Text.}
Continuing from the tabular checkpoint drives the batch reward from ~0.4 to ~0.6 on text examples. However, re-evaluating the new checkpoint on the \emph{tabular} test set
reveals measurable catastrophic forgetting.
The degradation happens across question types: Q3 drops from 0.326 to 0.264 ($-$0.06) and Q6 from 0.780 to 0.728 ($-$0.05).
Both Q-types demand accurate numerical reasoning over feature rankings or feature value intervals -- skills reinforced during tabular training and partially overwritten as the gradient signal shifts to text-span operations.

\paragraph{Phase 3 - Vision.}
Vision training for 200 steps shows a markedly weaker learning signal:
the batch reward rises only from 0.512 to 0.578, compared to the steep
trajectories in Phases 1 and 2 (Figure~\ref{fig:training_curves}, left).
We attribute the limited learning to two compounding effects. First, by Phase 3 the model's policy is already tightly peaked around tabular/text output formats, leaving insufficient entropy to explore the vision-specific bounding-box. Second, the flat reward curve indicates that rollout diversity has collapsed: all sampled trajectories produce similar rewards, so the
group-normalized advantage $\hat{A}_k \approx 0$ for most rollouts, suppressing the gradient signal.

\subsection*{Hybrid Training}
\begin{figure*}[h]
  \centering
  \includegraphics[width=\linewidth]{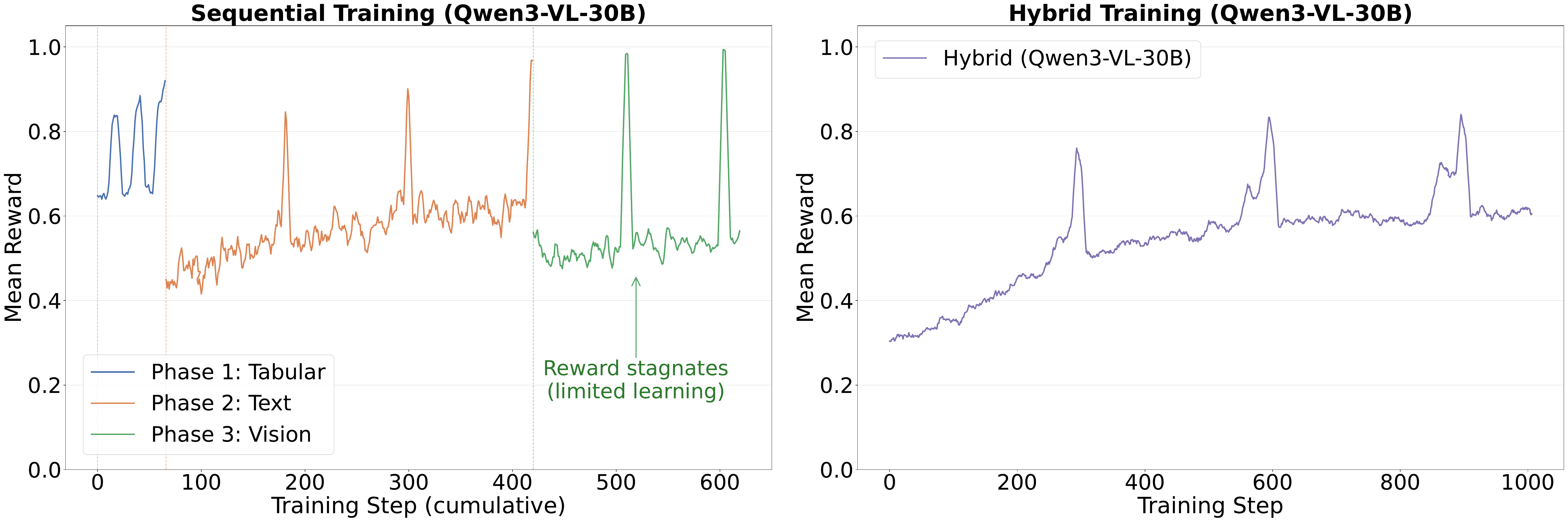}
  \caption{\xhdr{Training reward curves} \textbf{(Left)} \textit{Sequential training:} Vertical dashed lines mark phase boundaries (Tab $\to$ Text $\to$ Vision); the vision reward barely rises, indicating collapsed rollout diversity. \textbf{(Right)} \textit{Hybrid training (all modalities interleaved):} The recurring spikes are an artifact of the sampling seed, which concentrates relatively easy questions towards the end of each epoch.}
  \label{fig:training_curves}
\end{figure*}

Hybrid training on all three modalities simultaneously runs for 902 steps
across 3 epochs, with the batch reward rising from 0.313 to 0.594
(Figure~\ref{fig:training_curves}, right).
The slower convergence relative to any single-modality phase reflects the
wider diversity of the joint training distribution: the model must maintain
a flexible policy across tabular, text, and vision inputs simultaneously.

\subsection{Ablation Study: OOD Generalization}
\label{sec:ood_ablation}

We evaluate \method's out-of-distribution generalization along three axes: held-out question types within the proposed taxonomy, unseen datasets and model architectures, and a held-out question type entirely outside the taxonomy (concept-level explanation).

\xhdr{Held-out question types} We use spurious feature questions as OOD question types, as \method is never trained on them (see Sec.~\ref{sec:question_taxonomy}). Results in Table~\ref{tab:ood_q89_faithfulness} show that \method achieves higher overall faithfulness on tabular and text modalities, with particularly notable gains on Q8 for text. On vision inputs, the base agent slightly outperforms \method, suggesting that generalization to spurious feature reasoning remains challenging for vision modality without explicit training signal.

\begin{table}[h]
\centering
\small
\caption{OOD faithfulness scores on held-out question types Q8 and Q9. Best per row in \textbf{bold}.}
\label{tab:ood_q89_faithfulness}
\resizebox{0.63\columnwidth}{!}{%
\begin{tabular}{llcc}
\toprule
\textbf{Modality} & \textbf{Question Type} & \textbf{Base Agent} & \textbf{\method} \\
\midrule
\multirow{3}{*}{Tabular} & Q8 & 0.132 & \textbf{0.184} \\
& Q9 & \textbf{0.024} & 0.019 \\
\cline{2-4}
& \textit{Overall} & 0.081 & \textbf{0.106} \\
\midrule
\multirow{3}{*}{Text} & Q8 & 0.186 & \textbf{0.353} \\
& Q9 & 0.066 & \textbf{0.075} \\
\cline{2-4}
& \textit{Overall} & 0.116 & \textbf{0.186} \\
\midrule
\multirow{3}{*}{Vision} & Q8 & \textbf{0.068} & 0.050 \\
& Q9 & \textbf{0.070} & 0.023 \\
\cline{2-4}
& \textit{Overall} & \textbf{0.069} & 0.034 \\
\bottomrule
\end{tabular}}
\end{table}

\xhdr{Unseen datasets and model architectures.} Table~\ref{app:ood_data_faithfulness} reports faithfulness on entirely unseen dataset-model combinations for each modality. \method achieves the highest overall faithfulness across all three modalities, though it underperforms the base agent on text counterfactual and vision feature attribution, which we attribute to further hyperparameter tuning needed for better modality-specific trade-offs.

\begin{table}[h]
\centering
\small
\caption{OOD faithfulness scores on different datasets and model architectures. Best performance is \textbf{bolded}.}
\label{app:ood_data_faithfulness}
\resizebox{0.63\columnwidth}{!}{%
\begin{tabular}{llcc}
\toprule
\textbf{Modality} & \textbf{Question Group} & \textbf{Base Agent} & \textbf{\method} \\
\midrule
\multirow{4}{*}{Tabular} & Feature Attribution & 0.447 & \textbf{0.783} \\
& Counterfactual & 0.422 & \textbf{0.499} \\
& Spurious Features & 0.264 & \textbf{0.349} \\
\cline{2-4}
& \textit{Overall} & 0.379 & \textbf{0.581} \\
\midrule
\multirow{4}{*}{Text} & Feature Attribution & 0.428 & \textbf{0.563} \\
& Counterfactual & \textbf{0.860} & 0.770 \\
& Spurious Features & 0.639 & \textbf{0.778} \\
\cline{2-4}
& \textit{Overall} & 0.625 & \textbf{0.689} \\
\midrule
\multirow{4}{*}{Vision} & Feature Attribution & \textbf{0.621} & 0.583 \\
& Counterfactual & 0.246 & \textbf{0.521} \\
& Spurious Features & 0.416 & \textbf{0.624} \\
\cline{2-4}
& \textit{Overall} & 0.448 & \textbf{0.576} \\
\bottomrule
\end{tabular}}
\end{table}

\xhdr{Concept-level explanation generalization} \method's faithfulness protocol (Sec.~\ref{sec:metrics}) is fundamentally tied to \emph{local input-region perturbation}: masking a feature subset, token span, or bounding box. To test generalization beyond input-perturbation and beyond the trained question taxonomy entirely, we introduce a held-out question type, \textbf{Q11} (``Which concept was most responsible for the model's prediction?''), on CUB-200-2011, which asks the agent to name a semantic concept (\eg\ ``bill shape: hooked seabird'') rather than a region.

Faithfulness is scored by training a \textbf{Concept Activation Vector} (CAV)~\citep{kim2018tcav}, which is a linear probe on the model's penultimate-layer representation for each concept, and orthogonally projecting that direction out of the activation ($h' = h - (h \cdot \hat{v})\hat{v}$) before re-running only the classifier head; the probability drop $\max(0, P_{\mathrm{original}} - P_{\mathrm{modified}})$ is the faithfulness score, and the entire measurement happens in representation space. The Proposer/Actor pipeline still uses \method's existing tool library to decide which concept to name (Appendix~\ref{sec:xai_tools}); only the evaluation mechanism differs from Q1--Q10.

Table \ref{tab:q11_concept}shows that, across 120 questions (\texttt{cub\_resnet} + \texttt{cub\_densenet}), \method scores $\sim$18\% higher mean concept faithfulness than the base agent despite never being trained on this question type. Together, these results indicate \method's thought-action-observation loop generalizes across ablation targets (input region \emph{or} representation direction).

\begin{table}[h]
\centering
\small
\caption{Base agent vs.\ \method on Q11 (concept attribution), 120 questions.}
\label{tab:q11_concept}
\begin{tabular}{lcc}
\toprule
\textbf{Metric} & \textbf{Base Agent} & \textbf{\method} \\
\midrule
Mean faithfulness & 0.0316 & \textbf{0.0373} \\
Median faithfulness & 0.0150 & \textbf{0.0218} \\
Max faithfulness & 0.2398 & \textbf{0.2650} \\
Class-flip rate (ablation changes prediction) & 5/120 & 4/120 \\
\bottomrule
\end{tabular}
\end{table}

\subsection{Ablation Study: Prompt Robustness}
\label{sec:paraphrase}
To assess prompt robustness, we generate three semantically-equivalent paraphrases of every question in the taxonomy. Dynamic values (feature names, row IDs, class labels, masked tokens) are extracted
from the original question via regular expressions and re-injected verbatim into each paraphrase template, preserving semantic equivalence. Q1-Q4 receive three lexically distinct variants; Q5-Q7
and Q10 share a single uniform paraphrase across all three runs (variant-independent templates). Table~\ref{tab:paraphrase} lists the question-field phrasings for each type, illustrated with tabular-modality examples. Due to computational constraints, paraphrase evaluation is conducted on one dataset per modality.

\begin{figure}[tp]
\centering
\includegraphics[width=0.9\linewidth]{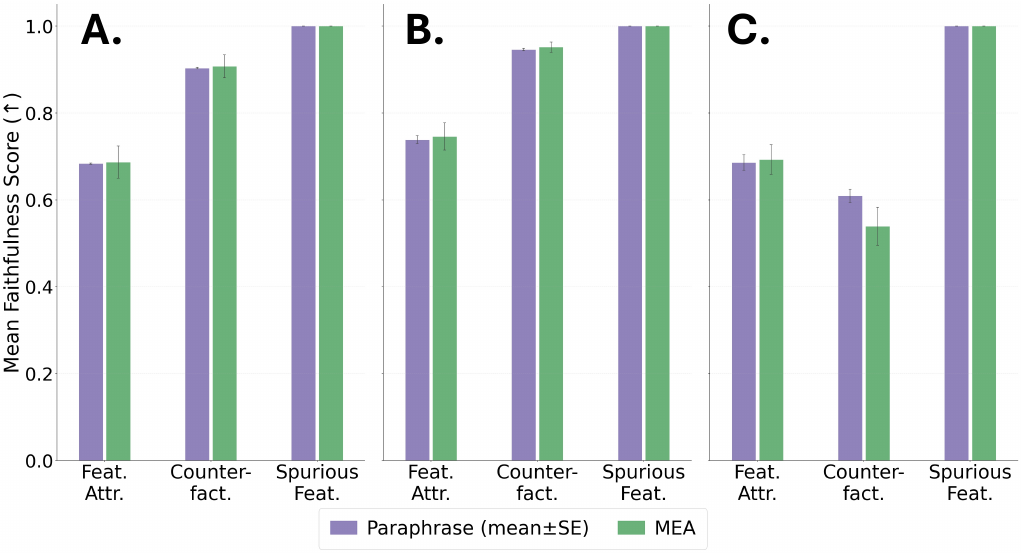}
\caption{\looseness=-1\xhdr{Ablation on prompt robustness} Mean faithfulness scores of \method evaluated on three paraphrased variants of the question taxonomy across \textbf{(A)} tabular, \textbf{(B)} text, and \textbf{(C)} vision modalities. Minimum variance indicates that \method is robust to changes in question phrasing.}
\label{fig:ablation_paraphrase}
\end{figure}

\begin{table*}[t]
\centering
\small
\caption{Three paraphrase variants of each question type (\textit{Q-field only};
tabular modality shown). Dynamic values such as feature names, row IDs, and class
labels (denoted \{\textit{placeholder}\}) are extracted from the original question
via regex and re-injected verbatim at generation time.}
\label{tab:paraphrase}
\setlength{\tabcolsep}{5pt}
\begin{tabular}{lp{3.6cm}p{3.6cm}p{3.6cm}}
\toprule
\textbf{Q-Type} & \textbf{Variant 1} & \textbf{Variant 2} & \textbf{Variant 3} \\
\midrule
Q1 & What was the most influential feature driving the model's prediction?
 & Which feature played the biggest role in this prediction?
 & What did the model rely on most heavily to reach this conclusion? \\
\addlinespace
Q2 & Which input feature had the least effect on the model's decision?
 & Which feature mattered least to the model's prediction?
 & Which feature is least relevant to the model's conclusion? \\
\addlinespace
Q3 & What features were most critical in setting this prediction apart from the
runner-up class?
 & Which features tipped the balance toward this prediction over the runner-up?
 & What distinguishes this prediction from the second most likely alternative? \\
\addlinespace
Q4 & What differences in the inputs account for the model giving instances A and B
different predictions?
 & Why does the model give different predictions to these two instances?
 & What input differences lead to contrasting predictions for these two
instances? \\
\addlinespace
Q5 & Would the model's output change if a specific feature were removed from this
input?
 & If a particular feature were excluded, would the model's prediction remain the
same?
 & How sensitive is the model's output to the presence of a specific feature? \\
\addlinespace
Q6 & What modifications to this instance would cause the model to output a
different class?
 & How should this input be changed to flip the model's prediction to a different
class?
 & What changes to this instance would lead the model to predict a different
outcome? \\
\addlinespace
Q7 & How would the model's output be affected if \{\textit{feature}\} were altered
or removed?
 & What would happen to the model's prediction if \{\textit{feature}\} were
modified or excluded?
 & In what way would the model's output change if \{\textit{feature}\} were no
longer present? \\
\addlinespace
Q10 & What explains the model's success on \{\textit{correct instances}\} but its
failure on \{\textit{wrong instance}\}?
  & Why does the model correctly classify \{\textit{correct instances}\} but fail
on \{\textit{wrong instance}\}?
  & What distinguishes \{\textit{wrong instance}\} from \{\textit{correct
instances}\} such that the model predicts it incorrectly? \\
\bottomrule
\end{tabular}
\end{table*}

\subsection{Ablation Study: Critic Agent}
\label{sec:critic_agent}
\begin{figure}[tp]
\centering
\includegraphics[width=\linewidth]{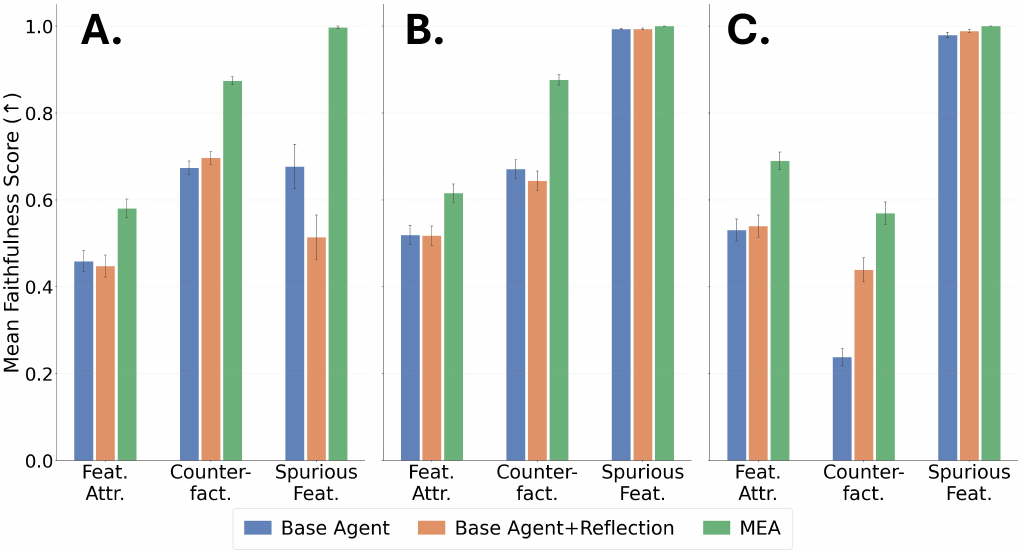}
\vspace{-0.075in}
\caption{\looseness=-1\xhdr{Ablation on critic agent} Mean faithfulness scores across three que1on categories for the base agent, base agent augmented with a critic agent, and \method, evaluated on \textbf{(A)} tabular, \textbf{(B)} text, and \textbf{(C)} vision modalities. \method consistently outperforms both baselines, demonstrating that RL-based training yields stronger improvements than inference-time critique.}
\vspace{-0.1in}
\label{fig:ablation_reflection}
\end{figure}

The Critic Agent augments the base agent pipeline with an inference-time reflection step. After the Actor Agent produces a strategy and explanation and receives a faithfulness score, the Critic receives a structured prompt and returns
two critiques: a \emph{Proposer Reflection} (feedback on tool selection) and an \emph{Actor Reflection} (feedback on explanation quality). The actor agent then regenerates its strategy and explanation conditioned on this feedback. Fig.~\ref{fig:ablation_reflection} shows that \method consistently outperforms Agent+Reflection despite the added inference-time step: on text counterfactual questions, Agent+Reflection scores $0.644$ while \method reaches $0.876$ (\textbf{+36\%}); on vision counterfactual questions, $0.439$ vs.\ $0.569$ (\textbf{+30\%}). Strikingly, on tabular spurious features the critic step even \emph{degrades} performance ($0.514$) relative to the base agent ($0.677$), while \method achieves $0.997$. These results demonstrate that RL-based end-to-end training provides substantially stronger and more reliable gains than inference-time critique.

\xhdr{Prompt}
Figure~\ref{fig:critic_prompt} shows the complete prompt structure passed to the Critic Agent. Each section is populated with runtime values extracted from the pipeline outputs.

\xhdr{Output Schema}
The Critc returns two feedbacks, separated by the markers \texttt{PROPOSER\_REFLECTION } and \texttt{ACTOR\_REFLECTION}. Table~\ref{tab:critic_output} describes each field. The Proposer Reflection is
consumed by the Proposer Agent to revise its tool selection; the Actor Reflection
is consumed by the Actor Agent to revise its explanation.

\begin{table*}[h]
\centering
\small
\caption{Output JSON schema of the critic agent.}
\label{tab:critic_output}
\setlength{\tabcolsep}{4pt}
\begin{tabular}{llp{6cm}}
\toprule
\textbf{Object} & \textbf{Field} & \textbf{Description} \\
\midrule
\multirow{6}{*}{\texttt{PROPOSER}}
& \texttt{analysis}              & Overall assessment of tool selection \\
& \texttt{tool\_analysis}        & Per-tool rating (\textit{high/medium/low}) and
recommendation (\textit{keep/remove/replace}) \\
& \texttt{tools\_to\_keep}       & Tools to retain in the revised strategy \\
& \texttt{tools\_to\_remove}     & Tools to drop \\
& \texttt{tools\_to\_add}        & New tools to include \\
& \texttt{strategy\_suggestions} & Free-text suggestions to improve tool
selection \\
\midrule
\multirow{5}{*}{\texttt{ACTOR}}
& \texttt{analysis}                  & Overall assessment of explanation quality
\\
& \texttt{identified\_issues}        & Specific errors or weaknesses in the
explanation \\
& \texttt{region\_feedback}          & Feedback on the accuracy of the identified
region or feature \\
& \texttt{explanation\_feedback}     & Feedback on clarity and correctness \\
& \texttt{improvement\_suggestions}  & Free-text suggestions to improve the
explanation \\
\bottomrule
\end{tabular}
\end{table*}

\begin{figure*}[t]
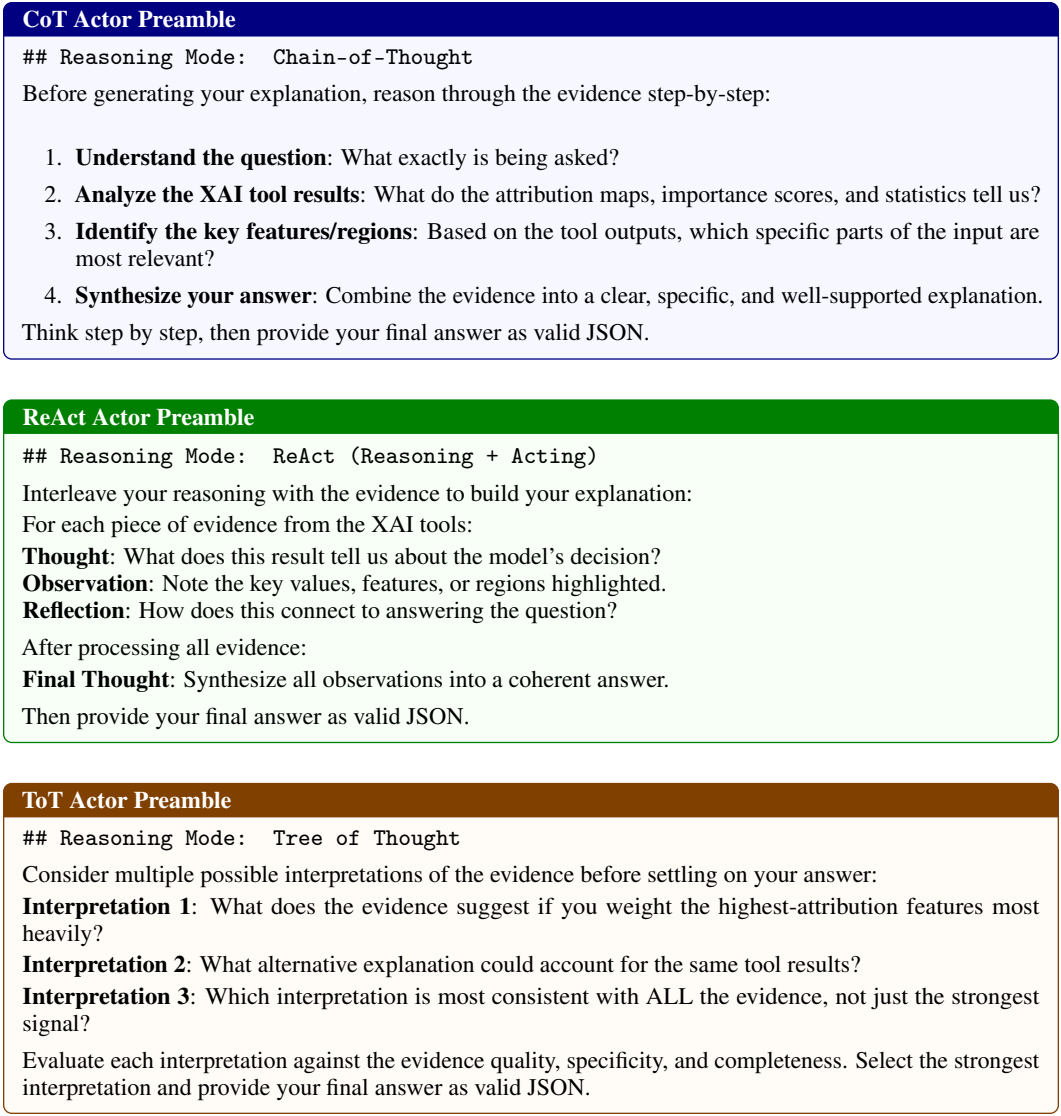

\centering

\begin{tcolorbox}[colback=blue!3, colframe=blue!50!black, title=\textbf{CoT Actor Preamble}, fonttitle=\small, boxrule=0.5pt, left=4pt, right=4pt, top=2pt, bottom=2pt]
\small\ttfamily
\#\# Reasoning Mode: Chain-of-Thought\\[4pt]
\rmfamily
Before generating your explanation, reason through the evidence step-by-step:\\[2pt]
\begin{enumerate}
    \item \textbf{Understand the question}: What exactly is being asked?
    \item \textbf{Analyze the XAI tool results}: What do the attribution maps, importance scores, and statistics tell us?
    \item \textbf{Identify the key features/regions}: Based on the tool outputs, which specific parts of the input are most relevant?
    \item \textbf{Synthesize your answer}: Combine the evidence into a clear, specific, and well-supported explanation.
\end{enumerate}
\vspace{2pt}
Think step by step, then provide your final answer as valid JSON.
\end{tcolorbox}

\vspace{4pt}

\begin{tcolorbox}[colback=green!3, colframe=green!50!black, title=\textbf{ReAct Actor Preamble}, fonttitle=\small, boxrule=0.5pt, left=4pt, right=4pt, top=2pt, bottom=2pt]
\small\ttfamily
\#\# Reasoning Mode: ReAct (Reasoning + Acting)\\[4pt]
\rmfamily
Interleave your reasoning with the evidence to build your explanation:\\[2pt]
For each piece of evidence from the XAI tools:\\[2pt]
\quad\textbf{Thought}: What does this result tell us about the model's decision?\\
\quad\textbf{Observation}: Note the key values, features, or regions highlighted.\\
\quad\textbf{Reflection}: How does this connect to answering the question?\\[4pt]
After processing all evidence:\\[2pt]
\quad\textbf{Final Thought}: Synthesize all observations into a coherent answer.\\[4pt]
Then provide your final answer as valid JSON.
\end{tcolorbox}

\vspace{4pt}

\begin{tcolorbox}[colback=orange!3, colframe=orange!50!black, title=\textbf{ToT Actor Preamble}, fonttitle=\small, boxrule=0.5pt, left=4pt, right=4pt, top=2pt, bottom=2pt]
\small\ttfamily
\#\# Reasoning Mode: Tree of Thought\\[4pt]
\rmfamily
Consider multiple possible interpretations of the evidence before settling on your answer:\\[2pt]
\textbf{Interpretation 1}: What does the evidence suggest if you weight the highest-attribution features most heavily?\\[2pt]
\textbf{Interpretation 2}: What alternative explanation could account for the same tool results?\\[2pt]
\textbf{Interpretation 3}: Which interpretation is most consistent with ALL the evidence, not just the strongest signal?\\[4pt]
Evaluate each interpretation against the evidence quality, specificity, and completeness. Select the strongest interpretation and provide your final answer as valid JSON.
\end{tcolorbox}

\caption{Generic actor preambles prepended to the actor prompt for each reasoning baseline across all question types.}
\label{fig:actor-preambles}
\end{figure*}

\subsection{Ablation Study: Proposer Strategy Mode}
\label{sec:proposer_mode_ablation}
The Proposer's strategy schema $S=(s_{\mathrm{type}},\mathcal{T}_{\mathrm{sel}},\mathcal{A}_{\mathrm{auto}})$ (Sec.~\ref{sec:actor_agent}) allows $s_{\mathrm{type}} \in \{\texttt{autonomous},\,\texttt{tools},\,\texttt{hybrid}\}$, but the main results always let the Proposer choose freely among the three. To isolate how much autonomous reasoning contributes beyond tool use alone, we force the Proposer into each single-mechanism regime -- \texttt{tool\_only} (external XAI tools alone, autonomous reasoning disabled) and \texttt{autonomous\_only} (self-reasoning alone, zero tool calls) -- for both the base agent and \method, over a fixed 500-question sample stratified by (dataset $\times$ question type) and allocated across modalities in proportion to test-set size (tabular 150 / text 195 / vision 155; Q1--Q7, Q10). Isolation is enforced at two levels: the disallowed branch's schema field is omitted from the Proposer prompt entirely, and a post-hoc check strips any leaked fields and fails the sample if the required branch is empty -- verified directly against execution logs (zero leaked autonomous tasks in \texttt{tool\_only} runs, zero tool invocations in \texttt{autonomous\_only} runs).

\begin{table}[h]
\centering
\small
\caption{Mean faithfulness by model and Proposer strategy mode.}
\label{tab:proposer_mode_overall}
\begin{tabular}{llrcc}
\toprule
\textbf{Model} & \textbf{Strategy Mode} & $n$ & \textbf{Mean Faith.} & \textbf{Pass Rate} \\
\midrule
\multirow{3}{*}{Base Agent} & Autonomous only & 493 & 0.515 & 48.5\% \\
& Tools only & 475 & 0.589 & 54.9\% \\
& Free choice (hybrid) & 2622 & 0.516 & 49.4\% \\
\midrule
\multirow{3}{*}{\method} & Autonomous only & 498 & 0.668 & 66.7\% \\
& Tools only & 498 & 0.731 & 74.7\% \\
& Free choice (hybrid) & 2654 & \textbf{0.748} & \textbf{76.5\%} \\
\bottomrule
\end{tabular}
\end{table}

Tool use is the stronger single mechanism for both models, but autonomous reasoning with \emph{zero} external tool calls already recovers 87.3\% (base) / 91.3\% (\method) of tool-only faithfulness. Training improves the autonomous-only arm \emph{more} than the tool-only arm ($+0.153$ vs.\ $+0.141$ mean faithfulness), narrowing the tool/autonomous gap after training. Most notably, for \method letting the Proposer choose freely ($0.748$) outperforms \emph{both} forced single-mechanism arms in every one of the three modalities (tabular $+0.006$, text $+0.018$, vision $+0.026$ over the better single mode), whereas for the base agent, free choice ($0.516$) is statistically indistinguishable from autonomous-only and clearly below tools-only ($0.589$) -- i.e., productively combining the two mechanisms is a capability training creates, not a free byproduct of offering both options. On Q6 (flip prediction) specifically, \method's free-choice arm ($0.822$) exceeds both tool\_only ($0.766$) and autonomous\_only ($0.782$) taken individually on the identical instances, a direct in-model demonstration that the two mechanisms are complementary rather than redundant.

\subsection{Ablation Study: Backbone Scale Transfer}
\label{sec:backbone_scale}
To test whether \method's training benefit transfers to a substantially smaller policy backbone, we repeat the Base-Agent-vs-\method comparison on \textbf{Qwen3.5-4B} \citep{qwen3.5} (roughly an order of magnitude smaller in active parameters), in addition to training-reward curves on a second backbone Qwen3-VL-30B-A3B-Instruct in Appendix~\ref{sec:training_strategy}. We train Qwen3.5-4B for a single epoch (300 LoRA steps, rank 32) with the identical reward pipeline, under the same evaluation protocol used throughout the paper.

Table~\ref{tab:4b_category} shows that, on this smaller 4B backbone, \method shows similar improvement as reported for the 30B \& 35B backbones, suggesting that training improves faithfulness in every modality overall (tabular $0.419\to0.671$ ($+60\%$), text $0.403\to0.575$ ($+43\%$), vision $0.142\to0.508$ ($+258\%$)), demonstrating that faithfulness gain comes from the training procedure rather than being an artifact specific to a single, large backbone.

\begin{table}[h]
\centering
\small
\caption{Faithfulness by question category, 4B backbone. Best per row in \textbf{bold}.}
\label{tab:4b_category}
\begin{tabular}{llcc}
\toprule
\textbf{Modality} & \textbf{Category} & \textbf{Baseline} & \textbf{Trained} \\
\midrule
\multirow{4}{*}{Tabular} & Feature Attribution & 0.446 & \textbf{0.578} \\
& Counterfactual & 0.398 & \textbf{0.763} \\
& Spurious Features & 0.356 & \textbf{0.933} \\
& \textit{Overall} & 0.419 & \textbf{0.671} \\
\midrule
\multirow{4}{*}{Text} & Feature Attribution & 0.325 & \textbf{0.534} \\
& Counterfactual & 0.413 & \textbf{0.480} \\
& Spurious Features & 0.660 & \textbf{0.960} \\
& \textit{Overall} & 0.403 & \textbf{0.575} \\
\midrule
\multirow{4}{*}{Vision} & Feature Attribution & 0.297 & \textbf{0.522} \\
& Counterfactual & 0.028 & \textbf{0.488} \\
& Spurious Features & 1.000 & 0.955 \\
& \textit{Overall} & 0.142 & \textbf{0.508} \\
\bottomrule
\end{tabular}
\end{table}

\subsection{Sanity Check: Synthetic Shortcut on a Biased Classifier}
\label{sec:bias_sanity_check}
A concern with any LLM-based explanation agent is that it may be faithful to the target model only when the model's decisions already align with the LLM's own semantic priors, rather than to the target model's actual (possibly idiosyncratic) decision process. To test this directly, we trained a \texttt{TwoLayerNN} on Adult Census with an injected synthetic shortcut column, \texttt{verification\_flag}, $\sim$90\% correlated with the true label (noise rate $0.10$, seed $42$) -- a feature whose dominance a domain expert or LLM prior would not anticipate for an income-prediction task. We independently verified the injected bias before any agent evaluation: overall test accuracy $0.939$; accuracy on rows where the flag agrees with the true label ($89.7\%$ of rows) is $0.981$, dropping to $0.575$ on the $10.3\%$ of rows where they disagree, and on those disagreement rows the model follows the flag (not the truth) $42.5\%$ of the time; an ablation flip-rate test confirms \texttt{verification\_flag} is $\sim$20$\times$ more influential on the model's output than a comparable real feature (education-num flip rate $0.8\%$ vs.\ $16.1\%$).

\begin{table}[h]
\centering
\small
\caption{Faithfulness score by question category on the biased classifier, paired $t$-test by (q\_type, row\_no).}
\label{tab:bias_sanity_check}
\begin{tabular}{lcccc}
\toprule
\textbf{Category} & \textbf{Base} & \textbf{\method} & $\Delta$ & $p$ \\
\midrule
Feature Attribution (Q1--4) & 0.407 & \textbf{0.626} & $+0.219$ & $1.0\mathrm{e}{-6}$ \\
Counterfactual (Q5--7) & 0.653 & \textbf{0.941} & $+0.288$ & $1.2\mathrm{e}{-5}$ \\
Spurious Features (Q10) & 0.725 & \textbf{1.000} & $+0.275$ & $0.0022$ \\
\bottomrule
\end{tabular}
\end{table}

\method wins or ties on every one of the 8 question types tested (Q1--Q7, Q10), with 5 of 8 significant at $p<0.05$ (Table~\ref{tab:bias_sanity_check}). Beyond the aggregate score, we checked whether each agent actually \emph{names} the shortcut when it matters: on the 13/20 Q1 rows where \texttt{verification\_flag} is verifiably the top driver by occlusion importance, \method names it correctly in 12/13 vs.\ 9/13 for the base agent; on Q2 (``least responsible feature''), the base agent incorrectly calls \texttt{verification\_flag} least responsible on 10/20 rows where it is actually dominant -- a genuine faithfulness failure -- while \method does this 0/20 times. On a classifier whose decision logic is engineered to deviate from human/LLM-intuitive reasoning, \method's explanations remain grounded in the model's actual (biased) decision process rather than reverting to prior-knowledge-consistent reasoning. We scope this finding to the single bias-injection design and tabular modality tested here, and note it does not directly test vision-specific shortcuts (\eg\ watermarks) or amplified real sensitive attributes.

\subsection{Statistical Significance Tests}
\label{sec:app_significance}
We report two-sided Welch's $t$-tests (unequal variance, unpaired) \cite{10.1093/biomet/34.1-2.28} for
comparisons against independent baselines, and paired $t$-tests for the
paraphrase ablation (where scores are matched by sample ID across conditions).
The significance level is $\alpha = 0.05$; ``n.s.'' denotes $p > 0.05$.
All values are mean $\pm$ standard error of the mean (SEM).

\xhdr{Vision Feature-Attribution Tool-Only Baselines}
\label{app:sig:tool}

Table~\ref{tab:sig:tool} compares \method (Q1, vision; $n=96$,
mean$=0.499 \pm 0.029$) against six tool-only XAI baselines evaluated on the
same vision feature-attribution task (Sensitivity had no evaluation data and
is excluded).
\method is statistically equivalent to three gradient-based methods (IG, Guided
Backprop, SmoothGrad) and significantly outperforms three others (SHAP, LIME,
GradCAM).

\begin{table*}[h]
\centering
\caption{Welch's $t$-test: \method vs.\ tool-only XAI baselines
         (vision, feature attribution Q1).
         \method mean $= 0.499 \pm 0.029$ ($n=96$).}
\label{tab:sig:tool}
\resizebox{0.9\linewidth}{!}{%
\footnotesize
\begin{tabular}{lrrrrrr}
\toprule
Baseline        & $n$ & Mean $\pm$ SEM & $\Delta$ (\method$-$BL) & $t$ & $p$ & Sig.\ \\
\midrule
SHAP            & 59  & $0.018 \pm 0.007$ & $+0.481$ & $16.26$ & $<0.001$ & * \\
LIME            & 96  & $0.259 \pm 0.027$ & $+0.240$ & $ 6.09$ & $<0.001$ & * \\
\textbf{Integrated Gradients} & \textbf{96} & $\mathbf{0.481 \pm 0.029}$ & $\mathbf{+0.018}$ & $\mathbf{0.44}$ & $\mathbf{0.661}$ & \textbf{n.s.} \\
\textbf{Guided Backprop}      & \textbf{96} & $\mathbf{0.497 \pm 0.027}$ & $\mathbf{+0.003}$ & $\mathbf{0.07}$ & $\mathbf{0.944}$ & \textbf{n.s.} \\
\textbf{SmoothGrad}           & \textbf{96} & $\mathbf{0.496 \pm 0.030}$ & $\mathbf{+0.003}$ & $\mathbf{0.08}$ & $\mathbf{0.935}$ & \textbf{n.s.} \\
GradCAM         & 96  & $0.131 \pm 0.017$ & $+0.368$ & $10.95$ & $<0.001$ & * \\
\bottomrule
\end{tabular}}
\end{table*}

\xhdr{Spurious-Feature Faithfulness (Q10)}
Table~\ref{tab:sig:q10} reports Welch's $t$-test results for the
spurious-feature question type (Q10) across text and vision modalities.
\method achieves \textbf{ceiling performance} ($1.000 \pm 0.000$, $n=160$) on
both modalities--every sample is scored correctly.
All Welch's $t$-tests return $p < 0.05$; however, this should be interpreted
in light of the ceiling effect: when one group has zero variance, any
non-zero difference in means will be flagged as significant regardless of
its practical magnitude.
For the strongest baselines the actual performance gap is negligible--Gemini-3.1
reaches $0.996$ on text (gap: $+0.004$) and Base Agent reaches $0.981$ on
vision (gap: $+0.019$)--indicating that top-performing models already
saturate Q10 and \method's advantage is marginal.
The significant $p$-values for weaker baselines (e.g.\ Claude-Haiku-4.5:
$0.665$ text, $0.855$ vision) do reflect a meaningful improvement.
CoT, ToT, and ReAct baselines had no Q10 evaluation data available.

\begin{table*}[h]
\centering
\caption{Welch's $t$-test: \method ($1.000 \pm 0.000$, $n=160$) vs.\
         baselines on spurious-feature questions (Q10).}
\label{tab:sig:q10}
\resizebox{0.9\linewidth}{!}{%
\begin{tabular}{llrrrrrr}
\toprule
Modality & Baseline & $n$ & Mean $\pm$ SEM & $\Delta$ (\method$-$BL) & $t$ & $p$ & Sig.\ \\
\midrule
\multirow{4}{*}{Text}
 & Base Agent       & 156 & $0.949 \pm 0.017$ & $+0.051$ & $ 3.09$ & $0.002$ & * \\
 & GPT-5.4          & 157 & $0.733 \pm 0.035$ & $+0.268$ & $ 7.67$ & $<0.001$ & * \\
 & Claude-Haiku-4.5 & 158 & $0.665 \pm 0.037$ & $+0.335$ & $ 9.11$ & $<0.001$ & * \\
 & Gemini-3.1       & 160 & $0.996 \pm 0.001$ & $+0.004$ & $ 2.92$ & $0.004$ & * \\
\midrule
\multirow{4}{*}{Vision}
 & Base Agent       & 148 & $0.981 \pm 0.004$ & $+0.019$ & $ 5.15$ & $<0.001$ & * \\
 & GPT-5.4          & 156 & $0.921 \pm 0.006$ & $+0.079$ & $12.53$ & $<0.001$ & * \\
 & Claude-Haiku-4.5 & 160 & $0.855 \pm 0.011$ & $+0.145$ & $13.58$ & $<0.001$ & * \\
 & Gemini-3.1       & 159 & $0.979 \pm 0.003$ & $+0.021$ & $ 6.54$ & $<0.001$ & * \\
\bottomrule
\end{tabular}}
\end{table*}

\xhdr{Paraphrase Robustness Ablation} Table~\ref{tab:sig:para} presents paired $t$-tests comparing Paraphrase
(mean of three paraphrase runs, matched on common sample IDs) against \method.
In 8 of 9 question-type$\times$modality conditions, there is no statistically
significant difference, confirming that \method's performance is robust to prompt
paraphrasing.
The single significant difference (vision counterfactual, $p=0.026$) shows
Paraphrase outperforming \method by $0.075$, suggesting that this question type
may benefit from additional training diversity.

\begin{table*}[h]
\centering
\caption{Paired $t$-test: Paraphrase vs.\ \method (matched common samples).
         $\Delta = $ Paraphrase $-$ \method.}
\label{tab:sig:para}
\resizebox{\linewidth}{!}{%
\begin{tabular}{llrrrrrrl}
\toprule
Modality & Question Group & $n_\text{pairs}$
         & Paraphrase ($\mu \pm$ SEM) & \method ($\mu \pm$ SEM)
         & $\Delta$ & $t$ & $p$ & Sig.\ \\
\midrule
\multirow{3}{*}{Tabular}
 & Feature Attribution & 106 & $0.683 \pm 0.037$ & $0.687 \pm 0.037$ & $-0.004$ & $-0.63$ & $0.532$ & n.s.\ \\
 & Counterfactual      &  32 & $0.934 \pm 0.023$ & $0.939 \pm 0.021$ & $-0.005$ & $-1.98$ & $0.056$ & n.s.\ \\
 & Spurious Features   &  17 & $1.000 \pm 0.000$ & $1.000 \pm 0.000$ & $ 0.000$ & --     & --     & n.s.\ \\
\midrule
\multirow{3}{*}{Text}
 & Feature Attribution & 126 & $0.738 \pm 0.030$ & $0.746 \pm 0.031$ & $-0.008$ & $-0.83$ & $0.407$ & n.s.\ \\
 & Counterfactual      &  85 & $0.947 \pm 0.012$ & $0.952 \pm 0.012$ & $-0.005$ & $-0.73$ & $0.469$ & n.s.\ \\
 & Spurious Features   &  37 & $1.000 \pm 0.000$ & $1.000 \pm 0.000$ & $ 0.000$ & --     & --     & n.s.\ \\
\midrule
\multirow{3}{*}{Vision}
 & Feature Attribution &  96 & $0.686 \pm 0.034$ & $0.693 \pm 0.034$ & $-0.007$ & $-0.56$ & $0.575$ & n.s.\ \\
 & \textbf{Counterfactual}       &  \textbf{72} & $\mathbf{0.614 \pm 0.041}$ & $\mathbf{0.539 \pm 0.044}$ & $\mathbf{+0.075}$ & $\mathbf{2.27}$ & $\mathbf{0.026}$ & \textbf{*} \\
 & Spurious Features   &  40 & $1.000 \pm 0.000$ & $1.000 \pm 0.000$ & $ 0.000$ & --     & --     & n.s.\ \\
\bottomrule
\end{tabular}}
\small\textit{Note:} ``--'' denotes that both conditions achieve identical ceiling
performance (all scores $= 1.0$), making $t$-test inapplicable.
\end{table*}

\begin{figure*}[!htbp]
    \centering

    \begin{tcolorbox}[
        enhanced,
        colback=teal!2,
        colframe=teal!70!black,
        boxrule=1pt,
        arc=4pt,
        left=8pt,
        right=8pt,
        top=8pt,
        bottom=8pt,
        title=\textbf{Critic Agent Prompt Template},
        colbacktitle=teal!70!black,
        coltitle=white,
        fonttitle=\bfseries\sffamily,
        fontupper=\small\sffamily,
        drop shadow=teal!15
    ]

    \textbf{\textcolor{teal!80!black}{System Role:}}

    \begin{tcolorbox}[
        colback=white,
        colframe=teal!20,
        boxrule=0.5pt,
        arc=2pt,
        left=4pt,
        right=4pt,
        top=4pt,
        bottom=4pt
    ]
    You are a critical analyst evaluating an XAI agent's strategy and explanation.
    \end{tcolorbox}

    \vspace{2mm}

    \textbf{\textcolor{teal!80!black}{User Input:}}

    \vspace{1mm}

    \begin{minipage}[t]{0.48\linewidth}
        \begin{tcolorbox}[
            colback=white,
            colframe=teal!20,
            boxrule=0.5pt,
            arc=2pt,
            left=4pt,
            right=4pt,
            top=4pt,
            bottom=4pt
        ]

        \textbf{Original Question} \\
        \pvar{question text / example context}

        \vspace{2.5mm}\hrule\vspace{2.5mm}

        \textbf{Question Context} \\
        Question Type: Q\pvar{1-10} \\
        Modality: \pvar{tabular $\vert$ text $\vert$ vision}

        \vspace{2.5mm}\hrule\vspace{2.5mm}

        \textbf{Strategy Used} \\
        Strategy Type: \pvar{tool-based} \\
        Selected Tools: \pvar{list of tool names} \\
        Strategy Reasoning: \pvar{Proposer's reasoning for tool selection}

        \end{tcolorbox}
    \end{minipage}
    \hfill
    \begin{minipage}[t]{0.48\linewidth}
        \begin{tcolorbox}[
            colback=white,
            colframe=teal!20,
            boxrule=0.5pt,
            arc=2pt,
            left=4pt,
            right=4pt,
            top=4pt,
            bottom=4pt
        ]

        \textbf{Generated Explanation} \\

        Output: \pvar{Actor's structured output dict} \\
        Explanation: \pvar{natural language explanation} \\
        Confidence: \pvar{Actor's self-reported confidence}

        \vspace{2.5mm}\hrule\vspace{2.5mm}

        \textbf{Faithfulness Evaluation} \\
        Score: \pvar{faithfulness score} \\
        Metric: \pvar{metric name and formula} \\
        $P_{\text{original}}$: \pvar{value}
        \quad
        $P_{\text{modified}}$: \pvar{value} \\
        $\vdots$

        \end{tcolorbox}
    \end{minipage}

    \vspace{2mm}

    \begin{tcolorbox}[
        colback=teal!5,
        colframe=teal!40!black,
        boxrule=0.5pt,
        arc=2pt,
        left=4pt,
        right=4pt,
        top=4pt,
        bottom=4pt
    ]

    \textbf{Task Instruction:}

    Analyze the strategy and results, then provide
    \textbf{two separate JSON outputs}:

    \begin{enumerate}
        \setlength{\itemsep}{2pt}
        \setlength{\parskip}{0pt}

        \item \texttt{PROPOSER\_REFLECTION}: feedback on tool selection strategy.

        \item \texttt{ACTOR\_REFLECTION}: feedback on explanation generation.
    \end{enumerate}

    \textit{
    Consider what went wrong, which tools were most or least helpful,
    and how both the strategy and explanation could be improved.
    }

    \end{tcolorbox}

    \end{tcolorbox}

    \caption{Complete prompt structure of the Critic Agent.}
    \label{fig:critic_prompt}

\end{figure*}

\subsection{Computational Cost}
\label{sec:app_computation_cost}
\xhdr{Model architecture and parameters}
\method uses \textbf{Qwen3.6-35B-A3B}~\citep{qwen36_35b_a3b} as its backbone, a
Mixture-of-Experts multi-modal language model with \textbf{35 billion total parameters} and
approximately \textbf{3.5 billion parameters activated per forward pass} (A3B). During inference for evaluation, we serve the model via LoRA adapters with rank 32, adding
$< 1\%$ additional parameters relative to the frozen base. Baseline comparisons using the un-finetuned backbone employed
\texttt{Qwen3.6-35B-A3B}, while closed-source
frontier baselines (Gemini-3.1-flash, GPT-5.4, Claude-Haiku-4.5) were accessed via their
respective public APIs.

\paragraph{GRPO training.}
All reinforcement learning training was conducted through \textit{Tinker} ~\citep{tml2025tinker}, a cloud-based training service that manages GPU allocation and distributed rollout execution internally; per-GPU-hour accounting is therefore not directly available from our
training logs. The main hybrid training run
ran for $\sim$93 GRPO steps
with a group size of 128-256 rollouts per step processed per step, covering all three modalities jointly. Each step produced a mean episode length of $\approx 3.6$ turns and $\approx 393$ action tokens.

\xhdr{Post-hoc explainers}
All post-hoc explainers and faithfulness metric computation, and ablations were run on a NVIDIA A100 GPUs (1 GPU per job, 16 CPU cores, 30 GB system RAM). Jobs were parallelized across datasets and question types (up to 4 concurrent jobs per
modality). The full test-set evaluation for a single modality (3.5K instances across
4 dataset-model combinations and 10 question types) required approximately
8-12 A100 GPU-hours depending on modality complexity, with vision being the most expensive due to image preprocessing and SmoothGrad/GradCAM computation. Summing across
the three modalities, primary test evaluations consumed an estimated
30-40 A100 GPU-hours. Ablation runs (prompt robustness, critic agent,
OOD datasets) added approximately an additional 20-30 A100 GPU-hours.
Total estimated evaluation compute is therefore on the order of 50-70 A100
GPU-hours.

\subsection{Prompt Templates for Proposer and Actor Agent}
\label{sec:prompts}

\begin{figure*}[!htbp]
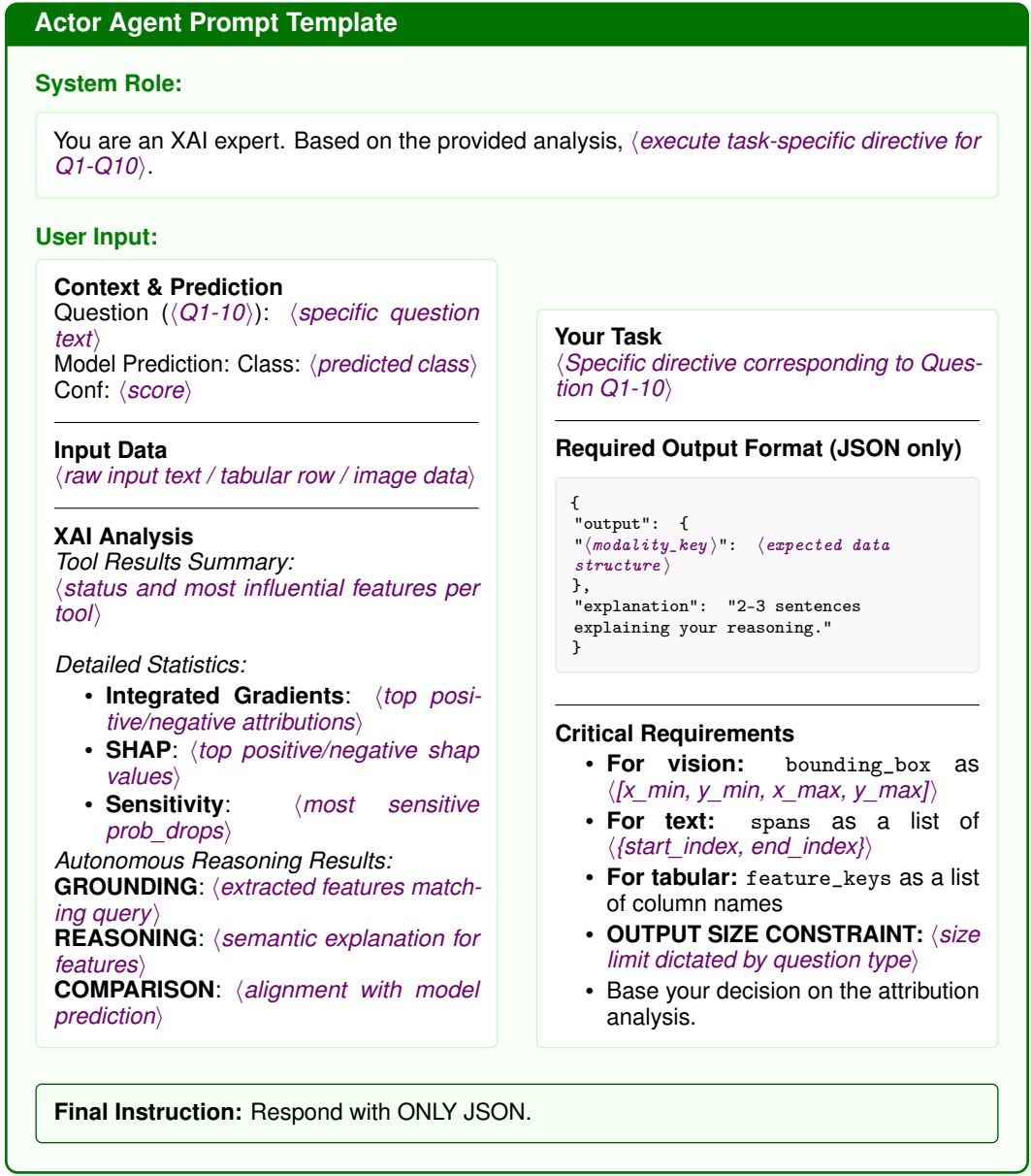

    \centering
    \begin{tcolorbox}[
        enhanced,
        colback=green!2,               
        colframe=green!45!black,       
        boxrule=1pt, arc=4pt,
        left=8pt, right=8pt, top=8pt, bottom=8pt,
        title=\textbf{Actor Agent Prompt Template},
        colbacktitle=green!45!black,
        coltitle=white,
        fonttitle=\bfseries\sffamily,
        fontupper=\small\sffamily,
        drop shadow=green!15           
    ]

    \textbf{\textcolor{green!50!black}{System Role:}}
    \begin{tcolorbox}[
        colback=white, colframe=green!25, boxrule=0.5pt, arc=2pt, 
        left=4pt, right=4pt, top=4pt, bottom=4pt
    ]
    You are an XAI expert. Based on the provided analysis, \pvar{execute task-specific directive for Q1-Q10}.
    \end{tcolorbox}

    \vspace{2mm}
    \textbf{\textcolor{green!50!black}{User Input:}}
    \vspace{1mm} 
    
    \begin{minipage}[t]{0.48\linewidth}
        \begin{tcolorbox}[
            colback=white, colframe=green!25, boxrule=0.5pt, arc=2pt, 
            left=4pt, right=4pt, top=4pt, bottom=4pt
        ]
        \textbf{Context \& Prediction} \\
        Question (\pvar{Q1-10}): \pvar{specific question text} \\
        Model Prediction: Class: \pvar{predicted class} \quad Conf: \pvar{score}
        
        \vspace{2.5mm}\hrule\vspace{2.5mm}
        
        \textbf{Input Data} \\
        \pvar{raw input text / tabular row / image data}
        
        \vspace{2.5mm}\hrule\vspace{2.5mm}
        
        \textbf{XAI Analysis} \\
        \textit{Tool Results Summary:} \\
        \pvar{status and most influential features per tool} \\
        
        \textit{Detailed Statistics:}
        \begin{itemize}
            \setlength{\itemsep}{1pt}\setlength{\parskip}{0pt}
            \item \textbf{Integrated Gradients}: \pvar{top positive/negative attributions}
            \item \textbf{SHAP}: \pvar{top positive/negative shap values}
            \item \textbf{Sensitivity}: \pvar{most sensitive prob\_drops}
        \end{itemize}
        
        \textit{Autonomous Reasoning Results:} \\
        \textbf{GROUNDING}: \pvar{extracted features matching query} \\
        \textbf{REASONING}: \pvar{semantic explanation for features} \\
        \textbf{COMPARISON}: \pvar{alignment with model prediction}
        \end{tcolorbox}
    \end{minipage}
    \hfill
    \begin{minipage}[t]{0.48\linewidth}
        \begin{tcolorbox}[
            colback=white, colframe=green!25, boxrule=0.5pt, arc=2pt,
            left=4pt, right=4pt, top=4pt, bottom=4pt
        ]
        \textbf{Your Task} \\
        \pvar{Specific directive corresponding to Question Q1-10}
    
        \vspace{2.5mm}\hrule\vspace{2.5mm}
        
        \textbf{Required Output Format (JSON only)} 
        \begin{tcolorbox}[
            colback=gray!5, colframe=gray!20, boxrule=0.5pt, arc=2pt,
            left=4pt, right=4pt, top=4pt, bottom=4pt, fontupper=\ttfamily\scriptsize
        ]
        \{ \\
        ~~"output": \{ \\
        ~~~~"\pvar{modality\_key}": \pvar{expected data structure} \\
        ~~\}, \\
        ~~"explanation": "2-3 sentences explaining your reasoning." \\
        \}
        \end{tcolorbox}
        
        \vspace{2.5mm}\hrule\vspace{2.5mm}
        
        \textbf{Critical Requirements} 
        \begin{itemize}
            \setlength{\itemsep}{2pt}\setlength{\parskip}{0pt}
            \item \textbf{For vision:} \texttt{bounding\_box} as \pvar{[x\_min, y\_min, x\_max, y\_max]}
            \item \textbf{For text:} \texttt{spans} as a list of \pvar{\{start\_index, end\_index\}}
            \item \textbf{For tabular:} \texttt{feature\_keys} as a list of column names
            \item \textbf{OUTPUT SIZE CONSTRAINT:} \pvar{size limit dictated by question type}
            \item Base your decision on the attribution analysis.
        \end{itemize}
        \end{tcolorbox}
    \end{minipage}

    \vspace{2mm}
    
    \begin{tcolorbox}[
        colback=green!5, colframe=green!35!black, boxrule=0.5pt, arc=2pt,
        left=4pt, right=4pt, top=4pt, bottom=4pt
    ]
    \textbf{Final Instruction:} Respond with ONLY JSON.
    \end{tcolorbox}

    \end{tcolorbox}
    \caption{Prompt structure for the Actor Agent. Task directives, constraints, and JSON schema keys are populated dynamically at runtime based on the specific question type (Q1-Q10) and data modality.}
    \label{fig:actor_prompt}
\end{figure*}

Refer to Fig.\ref{fig:actor_prompt} for the Actor Agent's prompt template and Fig.\ref{fig:proposer_prompt} for the Proposer Agent's prompt template.

\begin{figure*}[!htbp]
    \centering
    \begin{tcolorbox}[
        enhanced,
        colback=purple!2,               
        colframe=purple!50!black,       
        boxrule=1pt, arc=4pt,
        left=8pt, right=8pt, top=8pt, bottom=8pt,
        title=\textbf{Proposer Agent Prompt Template},
        colbacktitle=purple!50!black,
        coltitle=white,
        fonttitle=\bfseries\sffamily,
        fontupper=\small\sffamily,
        drop shadow=purple!15           
    ]

    \textbf{\textcolor{purple!60!black}{System Role:}}
    \begin{tcolorbox}[
        colback=white, colframe=purple!25, boxrule=0.5pt, arc=2pt, 
        left=4pt, right=4pt, top=4pt, bottom=4pt
    ]
    You are an AI explainability expert designing a strategy to answer a user's question about a machine learning model's prediction on \pvar{MODALITY}.
    \end{tcolorbox}

    \vspace{2mm}
    \textbf{\textcolor{purple!60!black}{User Input:}}
    \vspace{1mm} 
    
    \begin{minipage}[t]{0.48\linewidth}
        \begin{tcolorbox}[
            colback=white, colframe=purple!25, boxrule=0.5pt, arc=2pt, 
            left=4pt, right=4pt, top=4pt, bottom=4pt
        ]
        \textbf{Question \& Data Context} \\
        Question: \pvar{user question text} \\
        Input Content: \pvar{truncated input text, image, or tabular row}
        
        \vspace{2.5mm}\hrule\vspace{2.5mm}
        
        \textbf{Model Information} \\
        Model: \pvar{model name} \quad Arch: \pvar{architecture} \\
        Prediction: \pvar{class index \& name} (Conf: \pvar{score}) \\
        Top-K Predictions: \pvar{list of dicts containing top probabilities}
        
        \vspace{2.5mm}\hrule\vspace{2.5mm}
        
        \textbf{Available Methods} \\
        
        \textbf{1. Autonomous Analysis:} \pvar{instructions on using internal reasoning to ground/extract features} \\
        
        \textbf{2. External XAI Tools:}
        \begin{itemize}
            \setlength{\itemsep}{1pt}\setlength{\parskip}{0pt}
            \item \texttt{lime}: \pvar{executes local surrogate explanation}
            \item \texttt{integrated\_gradients}: \pvar{computes path attribution}
            \item \texttt{shap}: \pvar{computes shapley values}
            \item \texttt{sensitivity\_analysis}: \pvar{leave-one-out ablation}
            \item \texttt{guided\_backprop}: \pvar{modified ReLU gradients}
            \item \texttt{smoothgrad}: \pvar{averaged noisy saliency}
        \end{itemize}
        \end{tcolorbox}
    \end{minipage}
    \hfill
    \begin{minipage}[t]{0.48\linewidth}
        \begin{tcolorbox}[
            colback=white, colframe=purple!25, boxrule=0.5pt, arc=2pt,
            left=4pt, right=4pt, top=4pt, bottom=4pt
        ]
        \textbf{Your Task} \\
        \pvar{Specific strategy design directive corresponding to Question Q1-10}
    
        \vspace{2.5mm}\hrule\vspace{2.5mm}
        
        \textbf{Required Output Format (JSON only)} 
        \begin{tcolorbox}[
            colback=gray!5, colframe=gray!20, boxrule=0.5pt, arc=2pt,
            left=4pt, right=4pt, top=4pt, bottom=4pt, fontupper=\ttfamily\scriptsize
        ]
        \{ \\
        ~~"strategy\_type": "autonomous" | "tools" | "hybrid", \\
        ~~"reasoning": "Explain strategy choice (2-3 sentences)", \\
        ~~"autonomous\_tasks": [ \\
        ~~~~\{ \\
        ~~~~~~"task\_type": "grounding" | "reasoning" | "comparison", \\
        ~~~~~~"query": "Specific query for autonomous analysis", \\
        ~~~~~~"expected\_output": "What should be extracted" \\
        ~~~~\} \\
        ~~], \\
        ~~"tool\_selection": \{ \\
        ~~~~"selected\_tools": \pvar{subset of available tools}, \\
        ~~~~"reasoning": "Why these tools for this modality" \\
        ~~\} \\
        \}
        \end{tcolorbox}
        \end{tcolorbox}
    \end{minipage}

    \vspace{2mm}
    
    \begin{tcolorbox}[
        colback=purple!5, colframe=purple!35!black, boxrule=0.5pt, arc=2pt,
        left=4pt, right=4pt, top=4pt, bottom=4pt
    ]
    \textbf{Final Instruction:} Provide your strategy as a JSON object:
    \end{tcolorbox}

    \end{tcolorbox}
    \caption{Prompt structure for the Proposer Agent. Task directives are populated dynamically based on the active question type (Q1-Q10), ensuring the strategy generated aligns with the specific objective.}
    \label{fig:proposer_prompt}
\end{figure*}